\documentclass[letterpaper,conference]{IEEEtran} 
\usepackage[unicode]{hyperref}
\usepackage{xspace}
\usepackage{amsmath,amsthm,mathtools}
\usepackage[caption=false]{subfig}
\usepackage{flushend}
\usepackage[utf8x]{inputenc}

\newcommand{\DaDN}{\textit{DaDN}\xspace}
\newcommand{\BASE}{\textit{DaDN}\xspace}
\newcommand{\STRL}{\textit{Stripes}\xspace}
\newcommand{\STR}{\textit{STR}\xspace}
\newcommand{\TRTL}{\textit{Tartan}\xspace}
\newcommand{\TRT}{\textit{TRT}\xspace}

\newcommand{\tile}{tile\xspace}
\newcommand{\tiles}{tiles\xspace}

\newcommand{\fcl}{FCL\xspace}
\newcommand{\CVLs}{CVLs\xspace}
\newcommand{\FCLs}{FCLs\xspace}

\title{Tartan: Accelerating Fully-Connected and Convolutional Layers in Deep Learning Networks by Exploiting Numerical Precision Variability}

\author{Alberto Delmás Lascorz, Sayeh Sharify, Patrick Judd \& Andreas Moshovos\ \\
Electrical and Computer Engineering, University of Toronto\\
\texttt{\{delmasl1,sayeh,judd,moshovos\}@ece.utoronto.ca} \\
}

\begin{document}

\maketitle
\newcommand{\meanspeedupFC}{$1.61\times$\xspace}
\newcommand{\meanspeedupCV}{$1.91\times$\xspace}
\newcommand{\meanspeedupALL}{$1.90\times$\xspace} 

\newcommand{\meanefficiencyFC}{$1.06\times$\xspace}
\newcommand{\meanefficiencyCV}{$1.18\times$\xspace}
\newcommand{\meanefficiencyALL}{$1.17\times$\xspace} 

\newcommand{\meanspeedupFCloss}{$1.73\times$\xspace}
\newcommand{\meanspeedupCVloss}{$2.05\times$\xspace}
\newcommand{\meanspeedupALLloss}{$2.04\times$\xspace} 

\newcommand{\meanefficiencyFCloss}{$1.14\times$\xspace}
\newcommand{\meanefficiencyCVloss}{$1.26\times$\xspace}
\newcommand{\meanefficiencyALLloss}{$1.25\times$\xspace}  

\begin{abstract}
\TRTL\ ({\TRT}), a hardware accelerator for inference with Deep Neural Networks (DNNs), is presented and evaluated on Convolutional Neural Networks. \TRT exploits the variable per layer precision requirements of DNNs to deliver execution time that is proportional to the precision $p$ in bits used per layer for convolutional and fully-connected layers. Prior art has demonstrated an accelerator with the same execution performance only for convolutional layers\cite{Stripes-MICRO,Stripes-CAL}. Experiments on image classification CNNs show that on average across all networks studied,  \TRT outperforms a state-of-the-art bit-parallel accelerator~\cite{DaDiannao} by \meanspeedupALL without any loss in accuracy while it is \meanefficiencyALL more energy efficient. \TRT requires no network retraining while it enables trading off accuracy for additional improvements in execution performance and energy efficiency. For example, if a 1\% relative loss in accuracy is acceptable, \TRT is on average \meanspeedupALLloss faster and \meanefficiencyALLloss more energy efficient than a conventional bit-parallel accelerator. A Tartan configuration that processes 2-bits at time, requires less area than the 1-bit configuration, improves efficiency to $1.24\times$ over the bit-parallel baseline while being 73\% faster for convolutional layers and 60\% faster for fully-connected layers is also presented.
\end{abstract}

\section{Introduction}
\label{sec:intro}
It is only recently that commodity computing hardware in the form of graphics processors  delivered the performance necessary for practical, large scale Deep Neural Network applications~\cite{Alexnet}. At the same time, the end of Dennard Scaling in semiconductor technology~\cite{darkSilicon} makes it difficult to deliver further advances in hardware performance using existing general purpose designs. It seems that further advances in DNN sophistication would have to rely mostly on algorithmic and in general innovations at the software level which can be helped by innovations in hardware design. Accordingly, hardware DNN accelerators have emerged. The DianNao accelerator family was the first to use a wide single-instruction single-data (SISD) architecture to process up to 4K operations in parallel on a single chip ~\cite{diannao, DaDiannao} outperforming graphics processors by two orders of magnitude. Development in hardware accelerators  has since proceeded in two directions: either toward more general purpose accelerators that can support more machine learning algorithms while keeping performance mostly on par with DaDianNao (\DaDN)~\cite{DaDiannao}, or toward further specialization on  specific layers or classes of DNNs with the goal of outperforming \DaDN in execution time and/or energy efficiency, e.g.,~\cite{EIEISCA16,albericio:cnvlutin,Stripes-MICRO,isscc_2016_chen_eyeriss,Reagen2016}. This work is along the second direction. While an as general purpose as possible DNN\ accelerator is desirable further improving performance and energy efficiency for specific machine learning algorithms will provides us with the additional experience that is needed for developing the next generation of more general purpose machine learning accelerators.  
Section~\ref{sec:related} reviews several other accelerator designs.

While \DaDN's functional units process 16-bit fixed-point values, DNNs exhibit varying precision requirements across and within layers, e.g.,~\cite{judd:reduced}. Accordingly, it is possible to use shorter, per layer representations for activations and/or weights. However, with existing bit-parallel functional units doing so does not translate into a performance nor an energy advantage as the values are expanded into the native hardware precision inside the unit. Some designs opt to hardwire the whole network on-chip by using tailored datapaths per layer, e.g.,~\cite{kim_x1000_2014}. Such hardwired implementations are of limited appeal for  many modern DNNs  whose footprint ranges several 10s or 100s of megabytes of weights and activations.  Accordingly, this work targets accelerators that can translate any precision reduction into performance and that do not require that the precisions are hardwired at implementation time.

This work presents \TRTL (\TRT), a massively parallel hardware accelerator whose execution time for fully-connected (\FCLs) and convolutional (\CVLs) layers scales with the precision $p$ used to represent the input values. \TRT uses hybrid bit-serial/bit-parallel functional units and exploits the abundant parallelism of typical DNN layers with the following  goals: 1) exceeding \DaDN's execution time performance and energy efficiency, 2)\ maintaining the same activation and weight memory interface and wire counts, 3) maintaining wide, highly efficient accesses to weight and activation memories. Ideally, \TRTL  improves execution time over \BASE by $\frac{16}{p}$ where $p$ is the precision used for the activations  in \CVLs and for the activations and weights in \FCLs. Every single bit of precision that can be eliminated ideally reduces execution time and increases energy efficiency. For example, decreasing precision from 13 to 12 bits in an \fcl can ideally boost the performance improvement over \DaDN \BASE to 33\% from 23\% respectively. \TRT builds upon the \STRL (\STR) accelerator~\cite{Stripes-CAL,Stripes-MICRO} which improves execution time and energy efficiency on \CVLs\ only. While \STR\ matches the performance of a bit-parallel accelerator on \FCLs its energy efficiency suffers considerably. \TRT\ improves  performance and energy efficiency over a bit-parallel accelerator for both \CVLs and \FCLs. 

This work evaluates \TRT on a set of convolutional neural networks (CNNs) for image classification. On average \TRT reduces inference time by {{\meanspeedupFC}}, {{\meanspeedupCV}} and {{\meanspeedupALL}} over \DaDN for the fully-connected, the convolutional, and all layers  respectively. Energy efficiency compared to \DaDN with \TRT is {{\meanefficiencyFC}}, {{\meanefficiencyCV}} and {{\meanefficiencyALL}} respectively.
 By comparison, efficiency with \STR compared to \BASE is $0.73\times$,  $1.21\times$ and $1.14\times$ respectively. Additionally, \TRT enables trading off accuracy for improving execution time and energy efficiency. For example, on average on \FCLs, accepting a 1\% loss in relative accuracy improves performance to {{\meanspeedupFCloss}} and energy efficiency to {{\meanefficiencyFCloss}} compared to \DaDN.

In detail this work makes the following contributions:

\begin{itemize}
\item Extends the \STR accelerator offering performance improvements on \FCLs. Not only \STR does not improve performance on \FCLs, but its energy efficiency suffers compared to \BASE. 

\item \TRT incorporates  cascading multiple serial inner-product (SIP) units improving utilization when the number or filters or the dimensions of the filters is not a multiple of the datapath lane count.

\item 
It uses the methodology of Judd \textit{et al., \cite{judd:reduced}} to determine per layer weight and activation precisions for the fully-connected layers of several modern image classification CNNs.

 \item It evaluates a configuration of \TRT which trades off some of the performance improvement for enhancing energy and area efficiency.
The evaluated configuration processes two activation bits per cycle and requires half the parallelism and the SIPs than the bit-serial \TRT configuration. 

\item Reports energy efficiency and area measurements derived from a layout of the \TRT accelerator demonstrating its benefits over the preciously proposed \STR and \BASE accelerators.

\end{itemize}
 
The rest of this document is organized as follows: Section~\ref{sec:motivation} motivates \TRT. Section~\ref{sec:simple} illustrates the key concepts behind \TRT via an example. Section~\ref{sec:tartan} reviews the \DaDN architecture and presents an equivalent \TRTL configuration.  Section~\ref{sec:evaluation} presents the experimental results. Section~\ref{sec:related} reviews related work and discusses the limitations of this study and the potential challenges with \TRT. Section~\ref{sec:theend} concludes.

\section{Motivation}
\label{sec:motivation}
This section motivates \TRT by showing that: 1)  the precisions needed for the \FCLs of several modern image classification CNNs are far below the fixed 16-bit precision used by \BASE, and 2) the energy efficiency of \STR is below that of \BASE for \FCLs. Combined these results motivate \TRT which improves performance and energy efficiency for both \FCLs\ and \CVLs compared to \BASE.
\subsection{Numerical Representation Requirements Analysis}
\label{sec:motivation:precisions}
The experiments of this section corroborate past results that the precisions needed vary per layer for several modern image classification CNNs and during inference.  The section also shows that there is significant potential to improve performance if it were possible to exploit per layer precisions even for the \FCLs.  The per layer precision profiles presented here were found via the methodology of Judd \textit{et al.}~\cite{judd:reduced}.  Caffe~\cite{caffe} was used to measure how reducing the precision of each \fcl affects the network's overall \textit{top-1} prediction accuracy over 5000 images. The network definitions and pre-trained synaptic weights are taken from the Caffe Model Zoo~\cite{model-zoo}. The networks are used as-is without retraining. Further reductions in precisions may be possible with retraining.
As Section~\ref{sec:simple} will explain, \TRT's performance on an $\fcl$ layer $L$ is bound by the maximum of the weight ($P_w^L$) and activation ($P_a^L$) precisions. Accordingly, precision exploration was limited to cases where both $P^L_w$ and $P^L_a$ are equal. The search procedure is a gradient descent where a given layer's precision is iteratively decremented one bit at a time, until the network's accuracy drops. For weights, the fixed-point numbers are set to represent values between -1 and +1. For activations, the number of fractional bits is fixed to a previously-determined value known not to hurt accuracy, as per Judd \textit{et al.}\cite{judd:reduced}. While both activations and weights use the same number of bits, their precisions and ranges differ. For \CVLs only the activation precision is adjusted as with the \TRT design there is no benefit in adjusting the weight precisions as well. Weights remain at 16-bits for \CVLs. While, reducing the weight precision for \CVLs can reduce their memory footprint~\cite{ProteusICS16}, an option we do not explore further in this work.

Table~\ref{tab:TRN_precisions} reports the resulting per layer precisions separately for \FCLs and \CVLs. The ideal speedup columns report the performance improvement that would be possible if execution time could be reduced proportionally with  precision compared to a 16-bit bit-parallel baseline. For the \FCLs, the precisions required range from 8 to 10 bits and the potential for performance improvement is $1.64\times$ on average and ranges from $1.63\times$ to $1.66\times$. If a 1\% relative reduction in accuracy is acceptable then the performance improvement potential increases to $1.75\times$ on average and ranges from $1.63\times$ to as much as $1.85\times$. Given that the precision variability for \FCLs is relatively low (ranges from 8 to 11 bits) one may be tempted to conclude that a bit-parallel architecture with 11 bits may be an appropriate compromise. However, note that the precision variability is much larger for the \CVLs (range is 5 to 13 bits) and thus performance with a fixed precision datapath would be far below the ideal. For example, speedup with a 13-bit datapath would be just $1.23\times$ vs. the $2\times$ that is be possible with an 8-bit precision. A key motivation for \TRT is that  its  incremental cost  over \STR that already supports variable per layer precisions for \CVLs is well justified given the benefits. Section~\ref{sec:evaluation} quantifies this cost and the resulting performance and energy benefits.%

\begin{table*}[!t]
    \centering
    \begin{tabular}{|l|p{4cm}|r|r|r|}
    \hline
     & \multicolumn{2}{c|}{\textbf{Convolutional layers}} 
    & \multicolumn{2}{c|}{\textbf{Fully-Connected layers}}  \\
           \cline{2-5}
   &  \textbf{Per Layer Activation} &\textbf{Ideal} & \textbf{Per Layer Activation and}  &\textbf{Ideal} \\
  \textbf{Network} 
  & \textbf{Precision in Bits}  
  &\textbf{Speedup} 
  & \textbf{Weight Precision in Bits}  
  &\textbf{Speedup} \\ \hline
    \multicolumn{5}{|c|}{\textbf{100\% Accuracy}}\\\hline
       \hline  
AlexNet          & 9-8-5-5-7                                          & 2.38 & 10-9-9 & 1.66 \\ \hline
VGG\_S           & 7-8-9-7-9                                          & 2.04 & 10-9-9 & 1.64  \\ \hline
VGG\_M           & 7-7-7-8-7                                          & 2.23 & 10-8-8 & 1.64  \\ \hline
VGG\_19          & 12-12-12-11-12-10-11-11-13-12-13-13-13-13-13-13    & 1.35 &  10-9-9 & 1.63 \\ \hline \hline
    \multicolumn{5}{|c|}{\textbf{99\% Accuracy}}\\\hline
        \hline  
AlexNet           & 9-7-4-5-7                                          & 2.58 & 9-8-8 & 1.85 \\ \hline
VGG\_S            & 7-8-9-7-9                                          & 2.04 & 9-9-8 & 1.79 \\ \hline
VGG\_M            & 6-8-7-7-7                                          & 2.34 & 9-8-8 & 1.80 \\ \hline
VGG\_19           & 9-9-9-8-12-10-10-12-13-\allowbreak11-12-13-13-13-13-13        & 1.57 & 10-9-8 & 1.63 \\ \hline

\end{tabular}
\caption{
Per layer synapse precision profiles needed to maintain the same accuracy as in the baseline. \textit{Ideal}: Potential speedup with bit-serial processing of activations over a 16-bit bit-parallel baseline.
}

\label{tab:TRN_precisions}
\end{table*}

\subsection{Energy Efficiency with \STRL}
\label{sec:motivation:ee}

\begin{figure}
\centering
\includegraphics[scale=0.25]{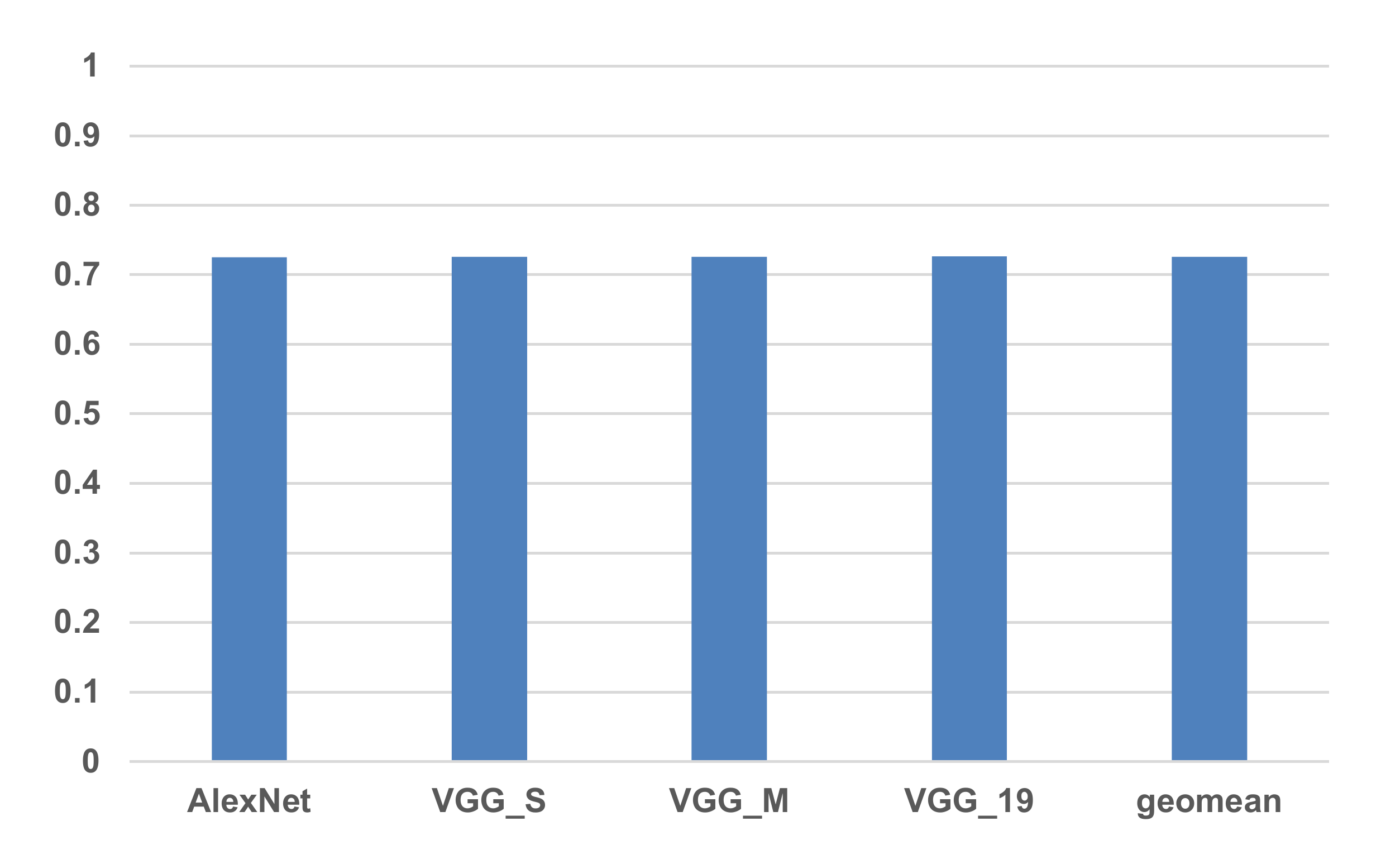}
\caption{Energy Efficiency of \STRL compared to \BASE on Fully-Connected layers.}
\label{fig:str-fcl-ee}
\end{figure}

\STRL (\STR) uses hybrid bit-serial/bit-parallel inner-product units for processing activations and weights respectively exploiting the per layer precision variability of modern CNNs~\cite{Stripes-MICRO}. However, \STR\ exploits precision reductions only for \CVLs\ as it relies on weight reuse across multiple windows to maintain the width of the weight memory the same as in \BASE\ (there is no weight reuse in \FCLs). Figure~\ref{fig:str-fcl-ee} reports the energy efficiency of \STR\ over that of \BASE for \FCLs (Section~\ref{sec:methodology} details the experimental methodology).
While performance is virtually identical to \BASE, energy efficiency is on average $0.73\times$ compared to \BASE. This result combined with the reduced precision requirements of \FCLs serves as motivation for extending \STR to improve performance and energy efficiency compared to \BASE\ on both \CVLs and \FCLs.

\subsection{Motivation Summary}
This section showed that: 1)~The per layer precisions for \FCLs on several modern CNNs for image classification vary significantly and exploiting them has the potential to improve performance by $1.64\times$ on average. 2) \STR that exploits variable precision requirements only for \CVLs achieves only $0.73\times$ the energy efficiency of a bit-parallel baseline. Accordingly, an architecture that would exploit precisions for \FCLs as well as \CVLs is worth investigating in hope that it will eliminate this energy efficiency deficit resulting in an accelerator that is higher performing and more energy efficient for both layer types. Combined \FCLs and \CVLs account for more than 99\% of the execution time in \BASE.

\section{\TRTL: A Simplified Example}
\label{sec:simple}
This section illustrates at a high-level the way \TRT operates by showing how it would process two purposely trivial cases: 1)~a fully-connected layer  with a single input activation producing two output activations, and 2)~a convolutional layer  with two input activations and one single-weight filter producing two output activations. The per layer calculations are:

\vspace{-15pt}
\begin{align*}
Fully-Connected: && Convolutional:\\
f_1 = w_1 \times a && c_1 = w \times a_1 \\
f_2 = w_2 \times a && c_2 = w \times a_2 \\
\end{align*}
\vspace{-25pt}

Where $f_1$, $f_2$, $c_1$ and $c_2$ are output activations, $w_1$, $w_2$, and $w$ are weights, and $a_1$, $a_2$ and $a$ are input activations. For clarity all values are assumed to be represented in 2 bits of precision. 

\subsection{Conventional Bit-Parallel Processing}
\begin{figure*}
\centering
\includegraphics[width=0.90\textwidth]{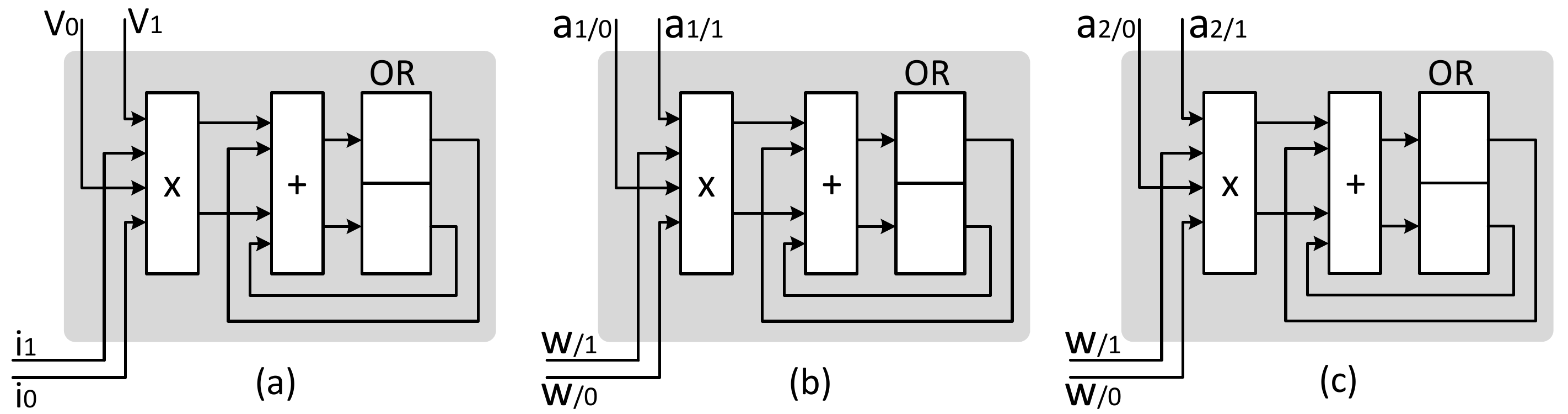}
\caption{Bit-Parallel Engine processing the convolutional layer over two cycles: a) Structure, b) Cycle 1, and c) Cycle 2.}
\label{fig:simple-base}
\end{figure*}

Figure~\ref{fig:simple-base}a shows a bit-parallel processing engine representative of \DaDN. Every cycle, the engine can calculate the product of two 2-bit inputs, $i$ (weight) and $v$ (activation) and accumulate or store it into the output register $OR$. Parts (b) and (c) of the figure show how this unit can calculate the example CVL over two cycles. In part (b) and during \textbf{cycle 1}, the unit accepts along the $v$ input bits 0 and 1 of $a_1$ (noted as $a_{1/0}$ and $a_{1/1}$ respectively on the figure), and along the $i$ input bits 0 and 1 of $w$ and produces both bits of output $c_1$. Similarly, during \textbf{cycle 2} (part (c)), the unit processes $a_2$ and $w$ to produce $c_2$. In total, over two cycles, the engine produced two $2b\times 2b$ products. Processing the example FCL also takes two cycles: In the first cycle, $w_1$ and $a$ produce $f_1$, and in the second cycle $w_2$ and $a$ produce $f_2$. This process is not shown in the interest of space.

\begin{figure*}[!htbp]
\centering
\subfloat[Engine Structure]{
\centering
\includegraphics[width=0.66\textwidth]{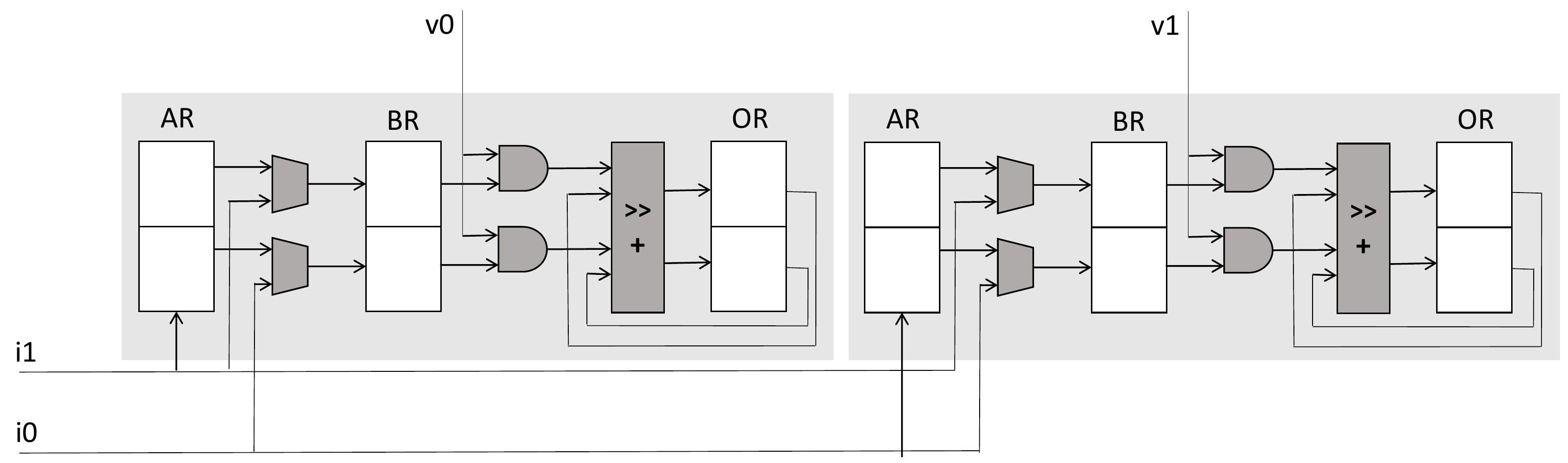}
\label{fig:simple-trt-cv-a}
}
\subfloat[Cycle 1: Parallel Load $w$ on BRs]{
\centering
\includegraphics[width=0.33\textwidth]{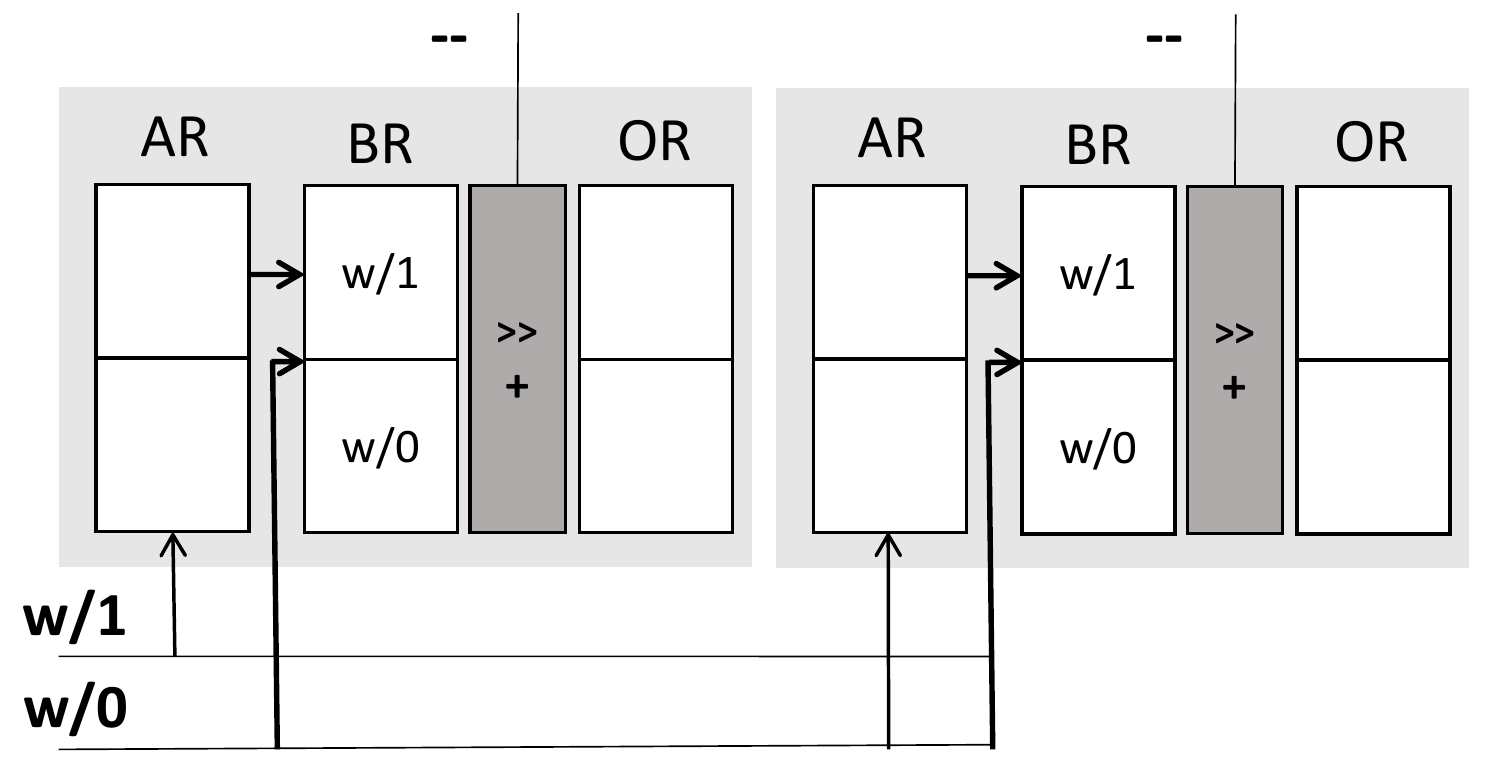}
\label{fig:simple-trt-cv-b}
}\\
\subfloat[Cycle 2: Multiply $w$ with bits 0 of the activations]{
\centering
\includegraphics[width=0.33\textwidth]{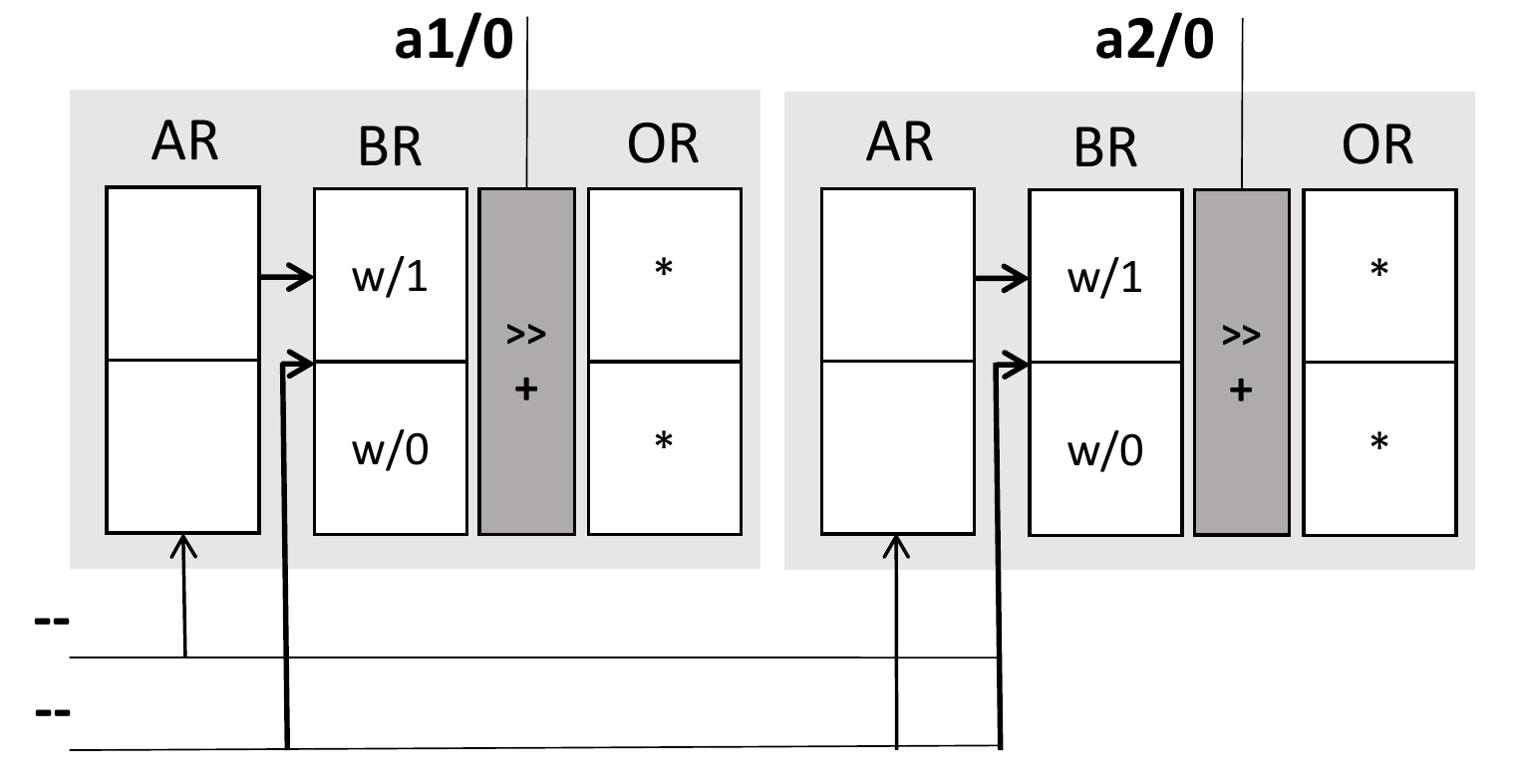}
\label{fig:simple-trt-cv-c}
}
\hspace{20pt}
\subfloat[Cycle 3: Multiply $w$ with bits 1 of the activations]{
\centering
\includegraphics[width=0.33\textwidth]{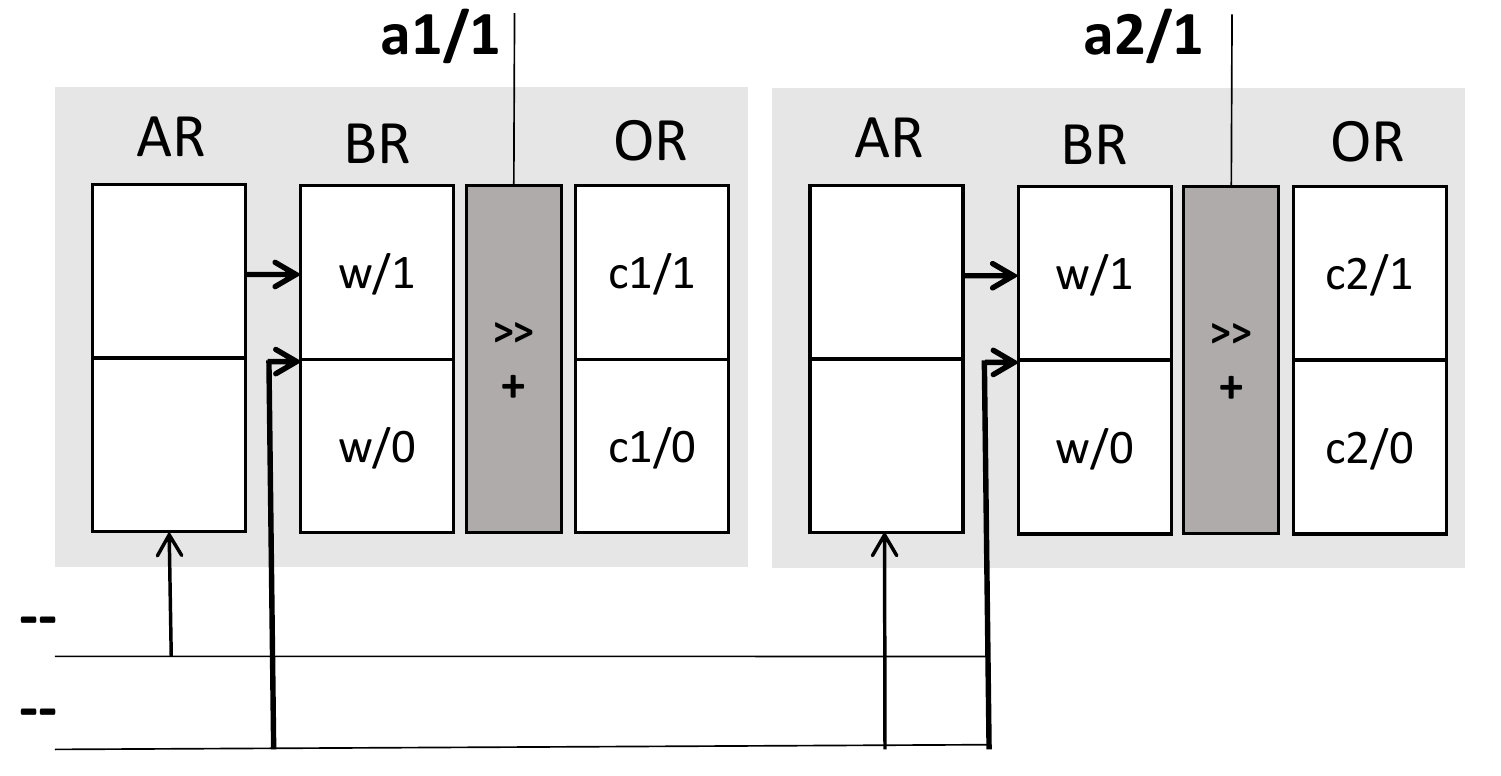}
\label{fig:simple-trt-cv-d}
}
\caption{Processing the example Convolutional Layer Using \TRT's Approach.
}
\label{fig:simple-trt}
\end{figure*}

\begin{figure*}[!htbp]
\centering
\subfloat[Cycle 1: Shift in bits 1 of weights into the ARs]{
\centering
\includegraphics[width=0.30\textwidth]{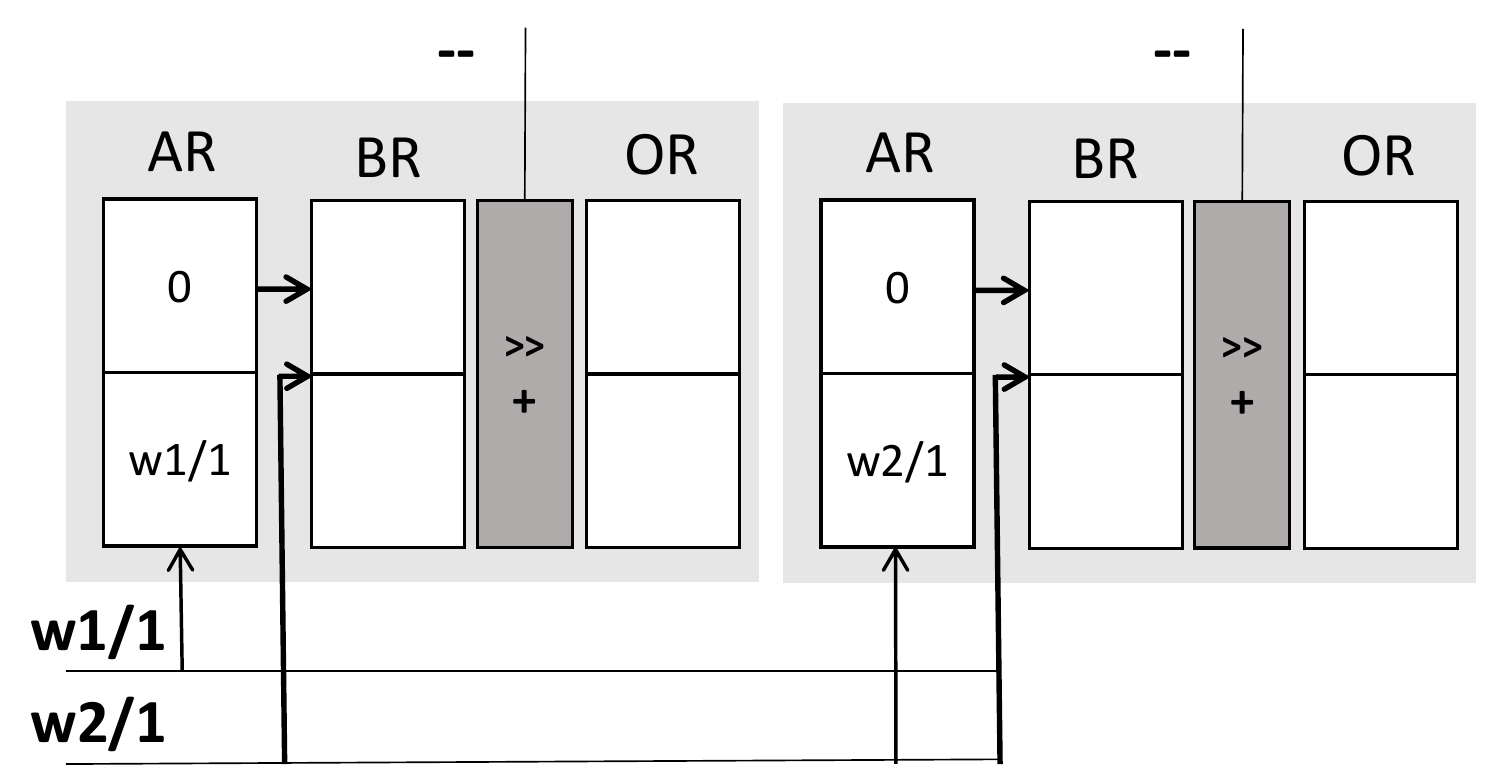}
\label{fig:simple-trt-fc-a}
}
\hspace{10pt}
\subfloat[Cycle 2: Shift in bits 0 of weights into the ARs]{
\centering
\includegraphics[width=0.30\textwidth]{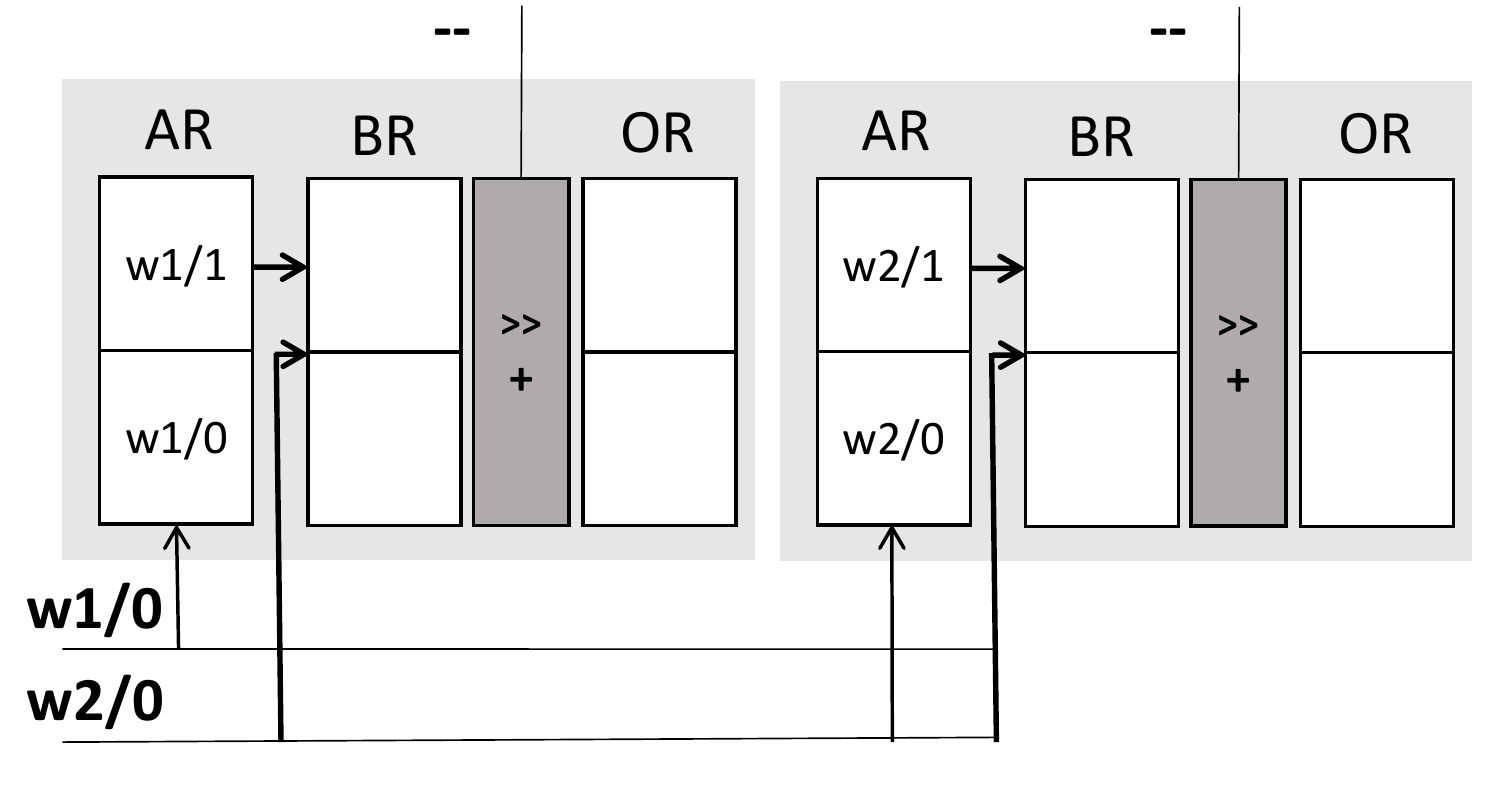}
\label{fig:simple-trt-fc-b}
}
\hspace{10pt}
\subfloat[Cycle 3: Copy AR into BR]{
\centering
\includegraphics[width=0.30\textwidth]{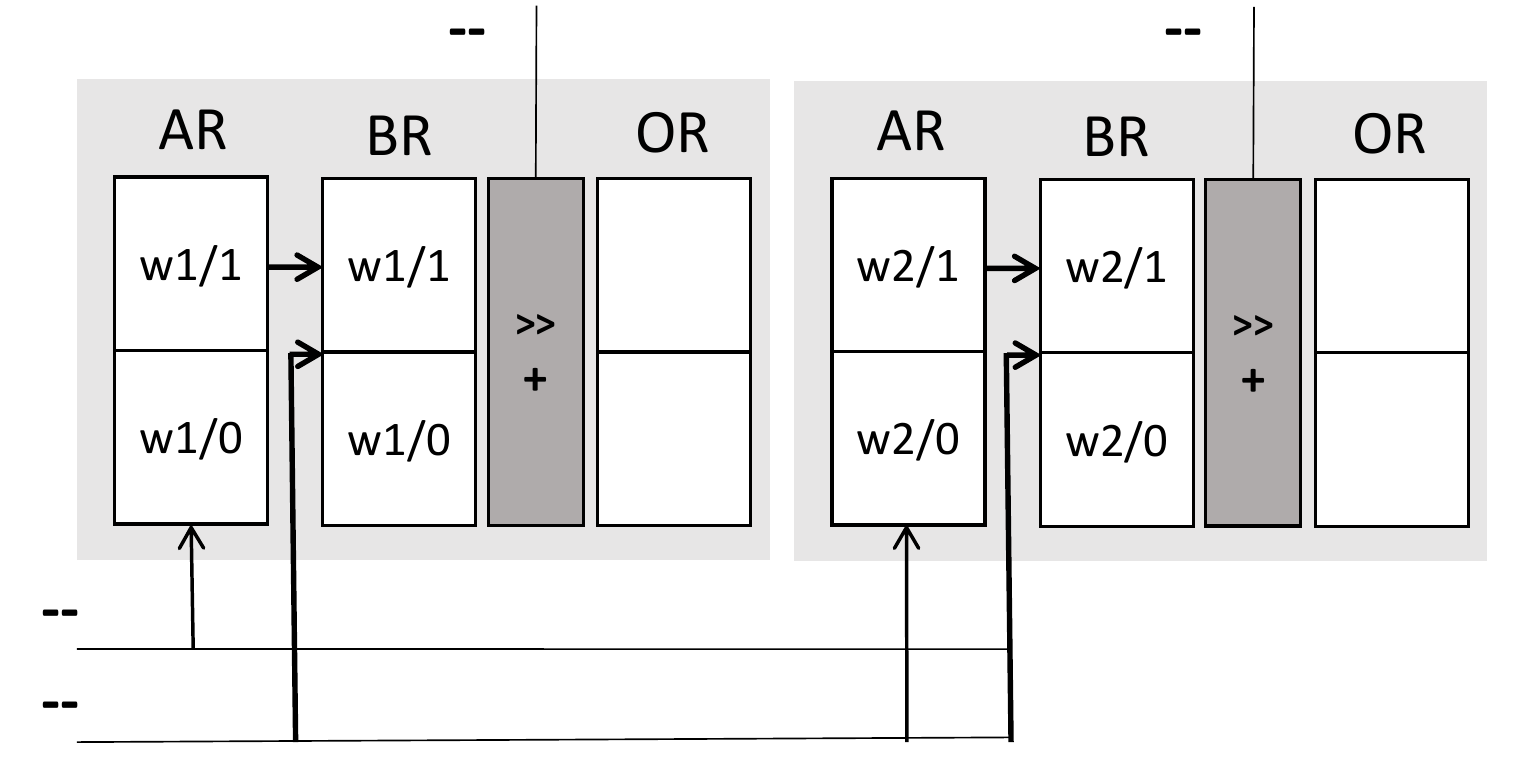}
\label{fig:simple-trt-fc-c}
}\\
\subfloat[Cycle 4: Multiply weights with first bit of $a$]{
\centering
\includegraphics[width=0.33\textwidth]{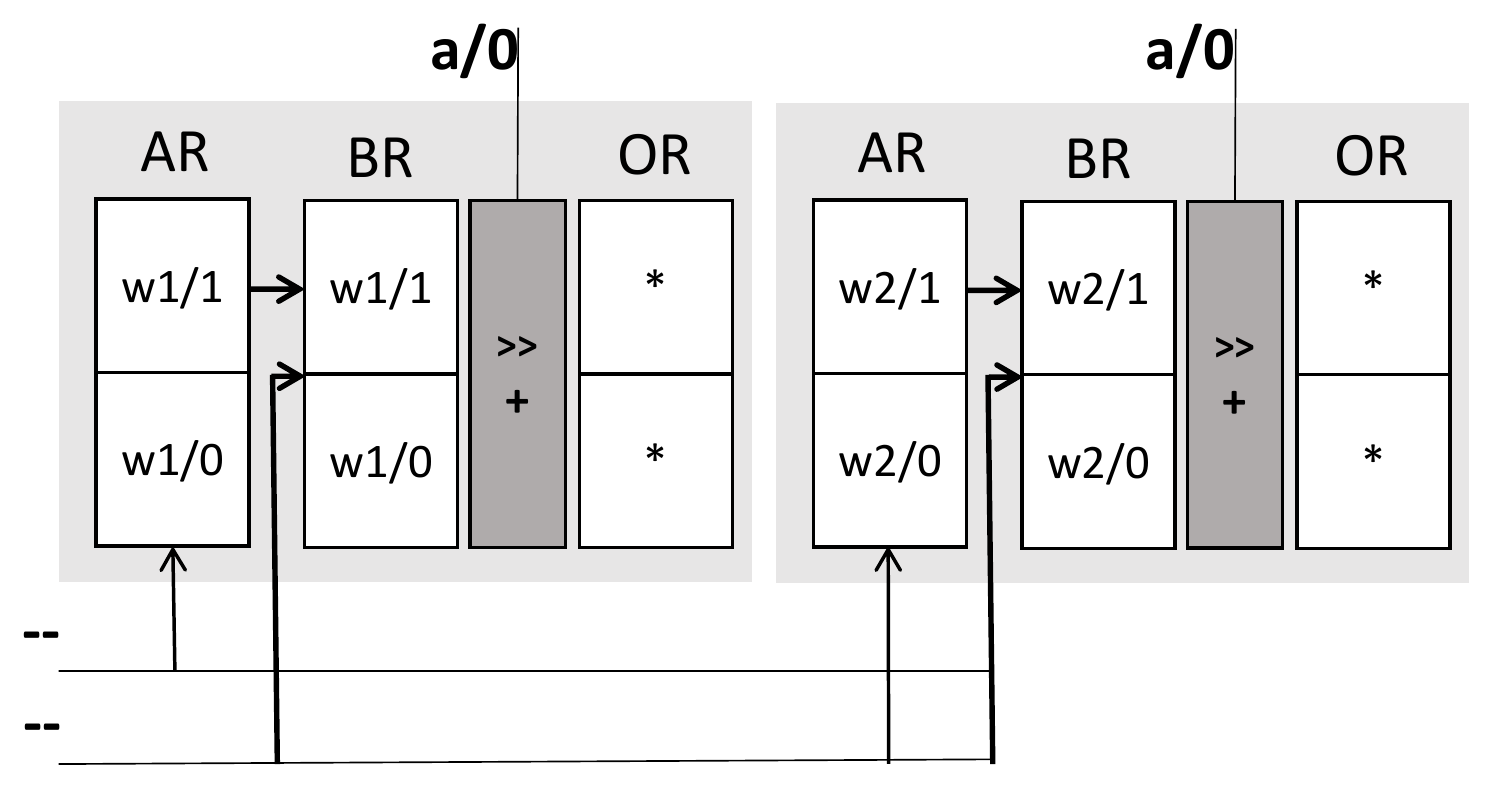}
\label{fig:simple-trt-fc-d}
}
\subfloat[Cycle 5: Multiply weights with second bit of $a$]{
\centering
\includegraphics[width=0.33\textwidth]{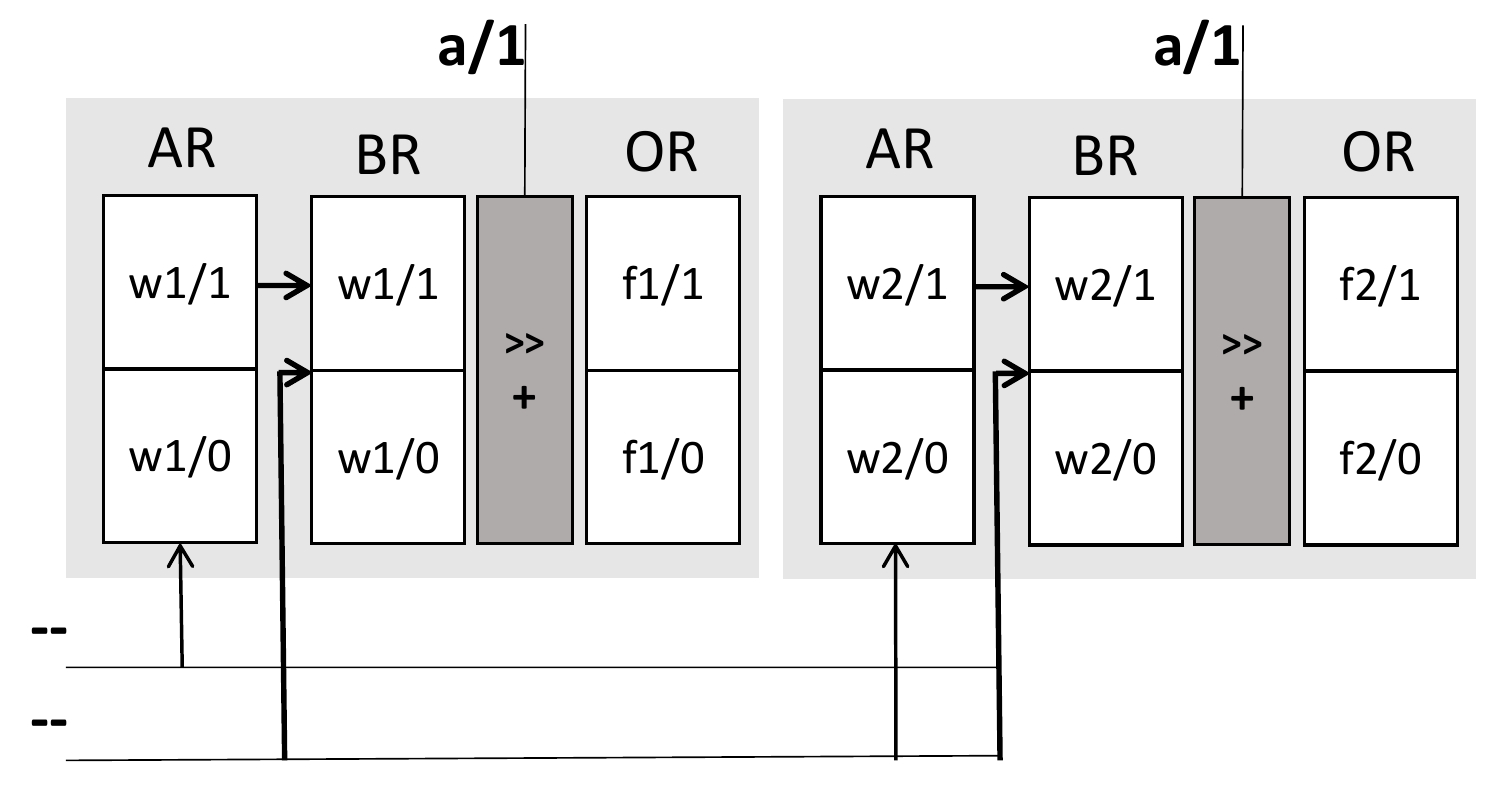}
\label{fig:simple-trt-fc-e}
}
\caption{Processing the example Fully-Connected Layer using \TRT's Approach.
}
\label{fig:simple-trt-fc}
\end{figure*}

\subsection{\TRTL's Approach}

Figure~\ref{fig:simple-trt} shows how a \TRT-like engine would process the example CVL. Figure~\ref{fig:simple-trt-cv-a} shows the engine's structure which comprises two subunits. The two subunits accept each one bit of an activation per cycle through inputs $v_0$ and $v_1$ respectively and as before, there is a common 2-bit weight input $(i_1,i_0)$. In total, the number of input bits is 4, the same as in the bit-parallel engine.

Each subunit contains three 2-bit registers: a shift-register AR, a parallel load register BR, and an parallel load output register OR. Each cycle each subunit can calculate the product of its single bit $v_i$ input with BR which it can write or accumulate into its OR.  There is no bit-parallel multiplier since the subunits process a single activation bit per cycle. Instead, two AND gates, a shift-and-add functional unit, and OR form a shift-and-add multiplier/accumulator. Each AR can load a single bit per cycle from one of the $i$ wires, and BR can be parallel-loaded from AR or from the $i$ wires.

\noindent\textbf{Convolutional Layer:}
Figure~\ref{fig:simple-trt-cv-b} through Figure~\ref{fig:simple-trt-cv-d} show how \TRT processes the CVL. The figures abstract away the unit details showing only the register contents. As Figure~\ref{fig:simple-trt-cv-b} shows, during \textbf{cycle 1}, the $w$ synapse is loaded in parallel to the BRs of both subunits via the $i_1$ and $i_0$ inputs. During \textbf{cycle 2}, bits 0 of $a_1$ and of $a_2$ are sent via the $v_0$ and $v_1$ inputs respectively to the first and second subunit. The subunits calculate concurrently $a_1/0\times w$ and $a_2/0\times w$ and accumulate these results into their ORs. Finally, in \textbf{cycle 3}, bit 1 of $a_1$ and $a_2$ appear respectively on $v_0$ and $v_1$. The subunits calculate respectively $a_1/1\times w$ and $a_2/1\times w$ accumulating the final output activations $c_1$ and $c_2$ into their ORs. 

In total, it took 3 cycles to process the layer. However, at the end of the third cycle, another $w$ could have been loaded into the BRs (the $i$ inputs are idle) allowing a new set of outputs to commence computation during cycle 4. That is, loading a new weight can be hidden during the processing of the current output activation for all but the first time. In the steady state, when the input activations are represented in two bits, this engine will be producing two $2b \times 2b$ terms every two cycles thus matching the bandwidth of the bit-parallel engine. 

If the activations $a_1$ and $a_2$ could be represented in just one bit, then this engine would be producing two output activations per cycle, twice the bandwidth of the bit-parallel engine. The latter is incapable of exploiting the reduced precision for reducing execution time. In general, if the bit-parallel hardware was using $P_{BASE}$ bits to represent the activations while only $P^{L}_{a}$ bits were enough, \TRT would outperform the bit-parallel engine by $\frac{P_{BASE}}{P_a^L}$.

\noindent\textbf{Fully-Connected Layer:} Figure~\ref{fig:simple-trt-fc} shows how a \TRT-like unit would process the example FCL. As Figure~\ref{fig:simple-trt-fc-a} shows, in \textbf{cycle 1}, bit 1 of $w_1$ and of $w_2$ appear respectively on lines $i_1$ and $i_0$. The left subunit's AR is connected to $i_1$ while the right subunit's AR is connected to $i_0$. The ARs shift in the corresponding bits into their least significant bit sign-extending to the vacant position (shown as a 0 bit on the example).
During \textbf{cycle 2}, as Figure~\ref{fig:simple-trt-fc-b} shows, bits 0 of $w_1$ and of $w_2$ appear on the respective $i$ lines and the respective ARs shift them in. At the end of the cycle, the left subunit's AR contains the full 2-bit $w_1$ and the right subunit's AR the full 2-bit $w_2$. 
In \textbf{cycle 3}, Figure~\ref{fig:simple-trt-fc-c} shows that each subunit copies the contents of AR into its BR. From the next cycle, calculating the products can now proceed similarly to what was done for the CVL. In this case, however, each BR contains a different weight whereas when processing the CVL in the previous section, all BRs held the same $w$ value. The shift capability of the ARs coupled with having each subunit connect to a different  $i$ wire allowed \TRT to load a different weight bit-serially over two cycles. Figure~\ref{fig:simple-trt-fc-d} and Figure~\ref{fig:simple-trt-fc-e} show cycles 4 and 5 respectively. During \textbf{cycle 4}, bit 0 of $a1$ appears on both $v$ inputs and is multiplied with the BR in each subunit. In \textbf{cycle 5}, bit 1 of $a1$ appears on both $v$ inputs and the subunits complete the calculation of $f_1$ and $f_2$. It takes two cycles to produce the two $2b \times 2b$ products once the correct inputs appear into the BRs. 

While in our example no additional inputs nor outputs are shown, it would have been possible to overlap the loading of a new set of $w$ inputs into the ARs while processing the current weights stored into the BRs. That is the loading into ARs, copying into BRs, and the bit-serial multiplication of the BRs with the activations is a 3-stage pipeline where each stage can take multiple cycles. In general, assuming that both activations and weights are represented using 2 bits, this engine would match the performance of the bit-parallel engine in the steady state. When both set of inputs $i$ and $v$ can be represented with fewer bits (1 in this example) the engine would produce two terms per cycle, twice the bandwidth of the bit-parallel engine of the previous section.

\noindent\textbf{Summary:} In general, if $P_{BASE}$ the precision of the bit-parallel engine, and $P_{a}^L$ and $P_{w}^L$ the precisions that can be used respectively for activations and weights for layer $L$, a \TRT engine can ideally outperform an equivalent bit parallel engine by $\frac{P_{BASE}}{P_{a}^L}$ for CVLs, and by $\frac{P_{BASE}}{max(P_{a}^L, P_{w}^L)}$ for FCLs. This example used the simplest \TRT engine configuration. Since typical layers exhibit massive parallelism, \TRT can be configured with many more subunits while exploiting weight reuse for CVLs and activation reuse for FCLs. The next section describes the baseline state-of-the-art DNNs accelerator and presents an equivalent \TRT configuration.

\section{\TRTL Architecture}
\label{sec:tartan}

This work presents \TRT as a modification 
of the state-of-the-art \textit{DaDianNao} accelerator. Accordingly, Section~\ref{sec:baseline} reviews \DaDN's design and how it can process FCLs and CVLs. For clarity, in what follows the term \textit{brick} refers to a set of 16 elements of a 3D activation or weight array input which are contiguous along the $i$ dimension, e.g., $a(x,y,i)...a(x,y,i+15)$. Bricks will be denoted by their origin element with a $B$ subscript, e.g., $a_B(x,y,i)$.  The size of a brick is a design parameter.
{Furthermore, an FCL can be thought of as a CVL where the input activation  array has unit x and y dimensions, and there are as many filters as output activations, and where the filter dimensions are identical to the input activation array.}%
\begin{figure*}
\centering
\subfloat[DaDianNao ]{
\centering
\includegraphics[width=0.40\textwidth]{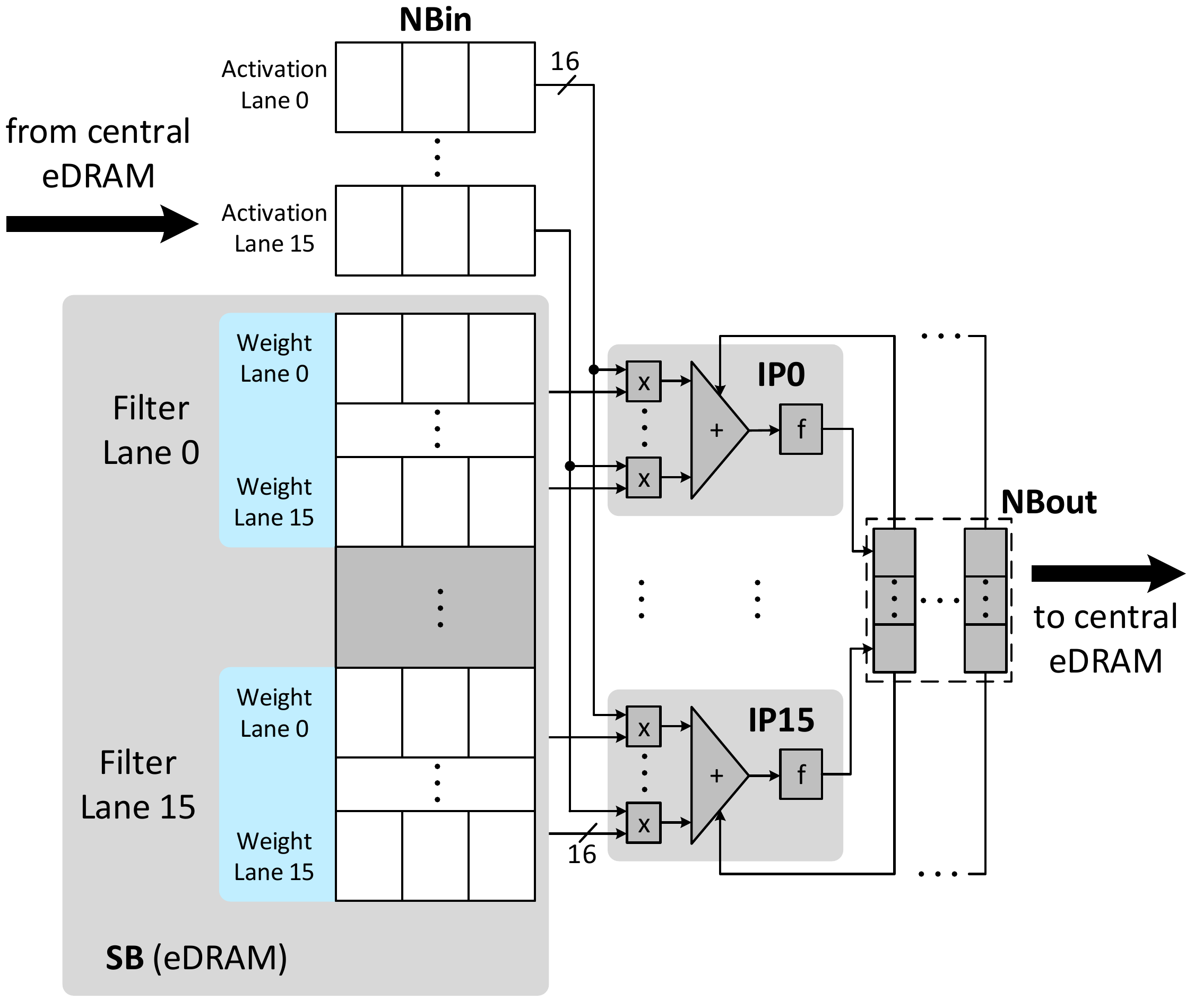}
\label{fig:dadiannao-overview}
}
\subfloat[\TRTL ]{
\centering
\includegraphics[width=0.6\textwidth]{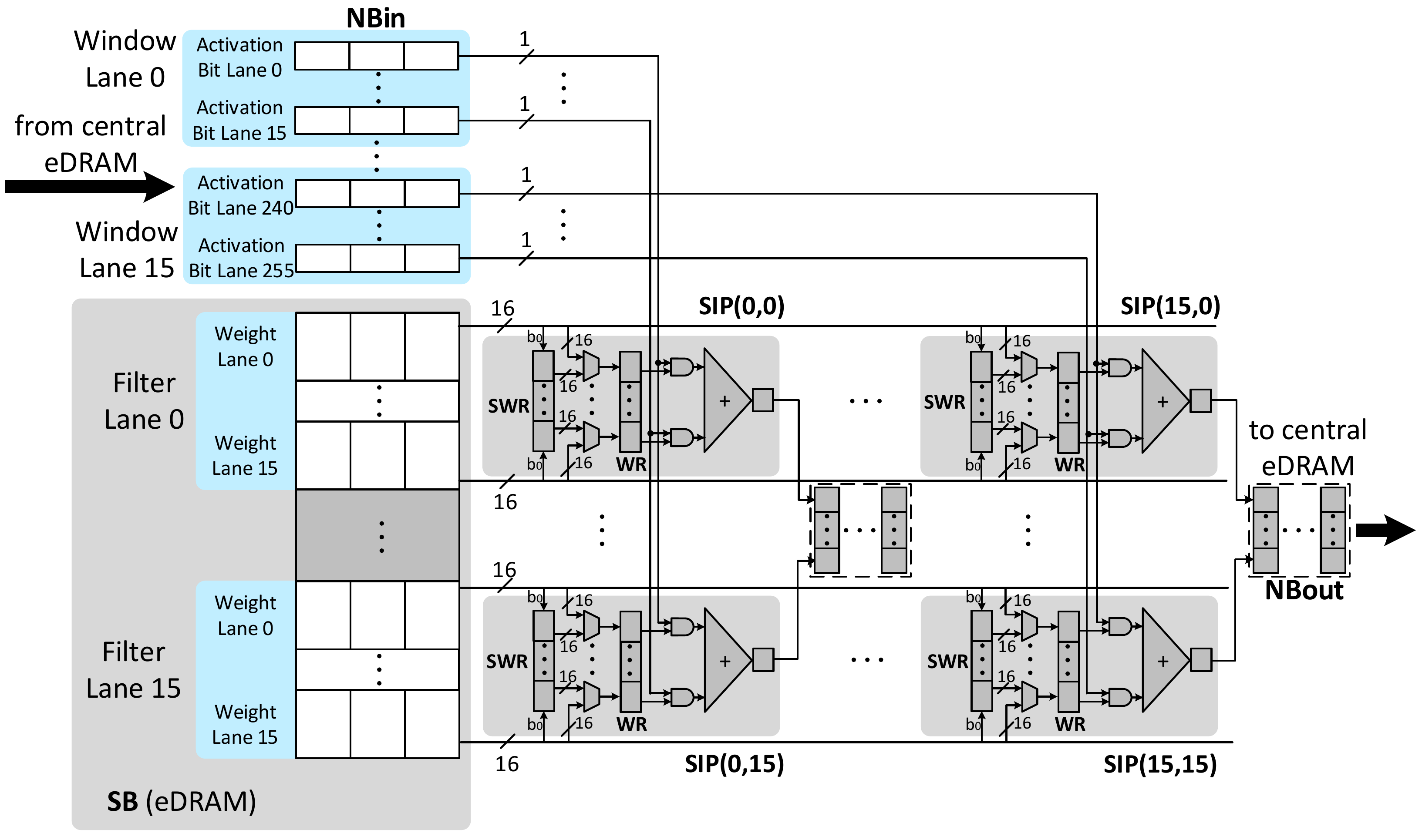}
\label{fig:tartan-overview}
}
\caption{Processing Tiles.} 
\end{figure*}

\begin{figure}[!t]
\centering
\includegraphics[scale=0.5]{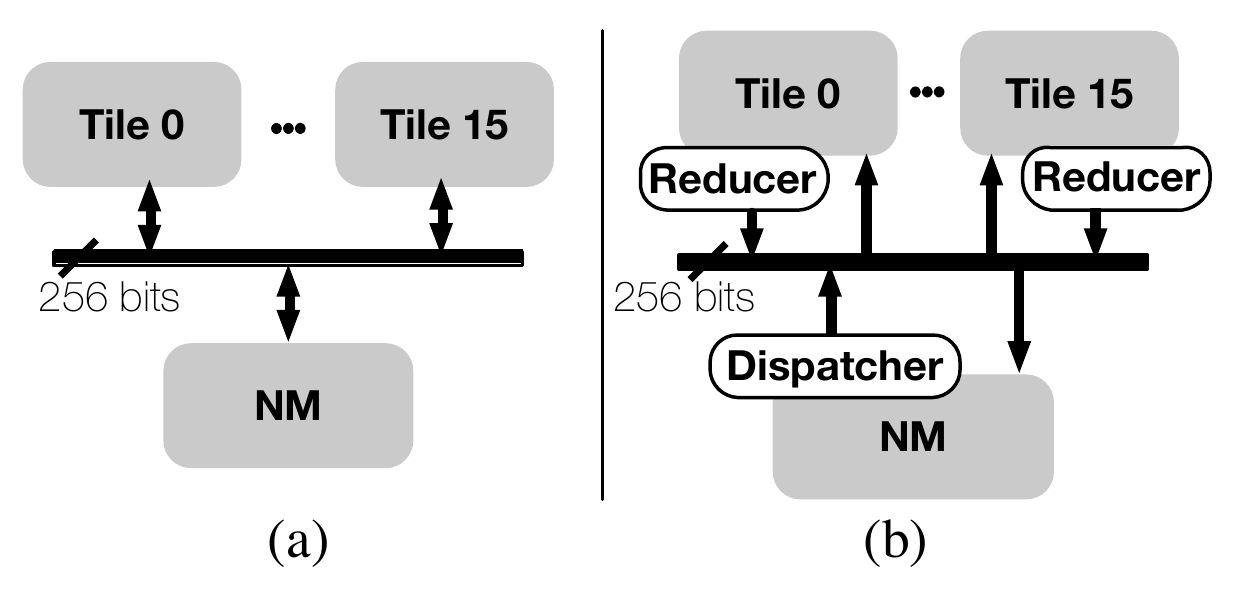}
\caption{
Overview of the system components and their communication. a) \BASE. b) \TRTL.
}
\label{fig:systems-overview}
\end{figure}

\subsection{Baseline System: DaDianNao}
\label{sec:baseline}

Figure~\ref{fig:dadiannao-overview} shows a \BASE \tile which processes 16 filters concurrently calculating 16 activation and weight products per filter for a total of 256 products per cycle~\cite{DaDiannao}. Each cycle the \tile accepts 16 weights per filter for total of 256 weight and 16 input activations. The \tile multiplies each weight with only one activation whereas each activation is multiplied with 16 weights, one per filter.  The \tile reduces the 16 products per filter into a single partial output activation, for a total of 16 partial output activations for the \tile. Each \BASE chip comprises 16 such \tiles, each processing a different set of 16 filters per cycle. Accordingly, each cycle, the whole chip processes 16 activations and $256\times 16=4K$ weights producing $16\times 16=256$ partial output activations, 16 per \tile.

Internally, each \tile has: 1)~a synapse buffer (SB) that provides 256 weights per cycle one per weight lane, 2)~an input neuron buffer (NBin) which provides 16 activations per cycle through 16 neuron lanes, and 3)~a neuron output buffer (NBout) which accepts 16 partial output activations per cycle. In the \tile's datapath each activation lane is paired with 16 weight lanes one from each filter. 
Each  weight and activation lane pair feeds a multiplier, and an  adder tree per filter lane reduces the 16 per filter 32-bit products into a partial sum. 
In all, the filter lanes produce each a partial sum per cycle, for a total of 16 partial output activations per tile.%
Once a full window is processed, the 16 resulting sums are fed through a non-linear activation function, $f$, to produce the 16 final output activations.  
The multiplications and reductions needed per cycle are implemented via 256 multipliers one per weight lane and sixteen 17-input (16 products plus the partial sum from NBout) 32-bit adder trees one per filter lane. 

Figure~\ref{fig:systems-overview}a shows an overview of the \BASE chip. There are 16 processing tiles connected via an interconnect to a shared 2MB central eDRAM \textit{Neuron Memory} (NM). \BASE's main goal was minimizing off-chip bandwidth while maximizing on-chip compute utilization. To avoid fetching weights from off-chip, \BASE uses a 2MB eDRAM Synapse Buffer (SB) for weights per \tile for a total of 32MB eDRAM for weight storage. All inter-layer activation outputs except for the initial input and the final output are stored in NM which is connected via a broadcast interconnect to the 16 Input Neuron Buffers (NBin) buffers. All values are 16-bit fixed-point, hence a 256-bit wide interconnect can broadcast a full activation brick in one step. Off-chip accesses are needed only for reading: 1)~the input image, 2)~the weights once per layer, and 3)~for writing the final output.

Processing starts by reading from external memory the first layer's filter weights, and the input image. The weights are distributed over the SBs and the input is stored into NM. Each cycle an input activation brick is broadcast to all units. Each units reads 16 weight bricks from its SB and produces a partial output activation brick which it stores in its NBout. Once computed, the output activations are stored through NBout to NM and then fed back through the NBins when processing the next layer. Loading the next set of weights from external memory can be overlapped with the processing of the current layer as necessary.  

\subsection{\TRTL}
\label{sec:design}

As Section~\ref{sec:simple} explained, \TRT processes activations bit-serially multiplying a single activation bit with a full weight per cycle. Each \BASE tile multiplies 16 16-bit activations with 256 weights each cycle. To match \BASE's computation bandwidth, \TRT needs to multiply 256 1-bit activations with 256 weights per cycle. Figure~\ref{fig:tartan-overview} shows the \TRT tile. It comprises 256 Serial Inner-Product Units (SIPs) organized in a $16\times 16$ grid. Similar to \DaDN each SIP  multiplies 16 weights with 16 activations and reduces these products into a partial output activation. Unlike \DaDN, each SIP accepts 16 single-bit activation inputs. Each SIP has two registers, each a vector of 16 16-bit subregisters: 1) the \textit{Serial Weight Register} (SWR), and 2) the \textit{Weight Register} (WR). These correspond to AR and BR of the example of Section~\ref{sec:simple}. NBout remains as in \BASE, however, it is distributed along the SIPs as shown.

\noindent\textbf{Convolutional Layers:} Processing starts by reading in parallel 256 weights from the SB as in \DaDN, and loading the 16 per SIP row weights in parallel to all SWRs in the row. Over the next $P_{a}^L$ cycles, the weights are multiplied by the bits of an input activation brick per column. \TRT exploits weight reuse across 16 windows sending a different input activation brick to each column. For example, for a CVL with a stride of 4 a \TRT \tile will processes 16 activation bricks $a_B(x,y,i)$, $a_B(x+4,y,i)$ through $a_B(x+63,y,i)$
in parallel a bit per cycle.  
Assuming that the \tile processes filters $f_{i}$ though $f_{i+15}$, after $P_{a}^L$ cycles it would produce the following 256 \textit{partial} output activations:
$o_B(x/4,y/4,f_{i})$, through $o_B(x/4+15,y/4,f_{i})$, 
that is 16 contiguous on the $x$ dimension output activation bricks. Whereas \BASE would process 16 activations bricks over 16 cycles, \TRT processes them concurrently but bit-serially over $P_{a}^L$ cycles. If $P_{a}^L$ is less than 16, \TRT will outperform \BASE by $16/P_{a}^L$, and when $P_{a}^L$ is 16, \TRT will match \BASE's performance. 

\noindent\textbf{Fully-Connected Layers:} Processing starts by loading bit-serially and in parallel over $P_{w}^L$ cycles, 4K weights into the 256 SWRs, 16 per SIP. Each SWR per row gets a different set of 16 weights as each subregister is connected to one out of the 256 wires of the SB output bus for the SIP row (is in \BASE\ there are  $256\times 16=4K$ wires).
Once the weights have been loaded, each SIP copies its SWR to its SW and multiplication with the input activations can then proceed bit-serially over $P_{a}^L$ cycles. Assuming that there are enough output activations so that a different output activation can be assigned to each SIP, the same input activation brick can be broadcast to all SIP columns. For example, for an FCL a \TRT \tile will process one activation brick $a_B(i)$ bit-serially to produce 16 output activation bricks $o_B(i)$ through $o_B(i\times 16)$ one per SIP column. Loading the next set of weights can be done in parallel with processing the current set, thus execution time is constrained by $P_{max}^L = max(P_{a}^L,P_{w}^L)$. Thus, a \TRT \tile produces 256 partial output activations every $P_{max}^L$ cycles, a speedup of $16/P_{max}$ over \BASE since a \BASE \tile always needs 16 cycles to do the same.

\noindent\textbf{Cascade Mode:} For \TRT to be fully utilized an FCL must have at least 4K output activations. Some of the networks studied have a layer with as little as 2K output activations. To avoid underutilization, the SIPs along each row are cascaded into a daisy-chain, where the output of one can feed into an input of the next via a multiplexer. This way, the computation of an output activation can be sliced over the SIPs along the same row. In this case, each SIP processes only a portion of the input activations resulting into several partial output activations along the SIPs on the same row. Over the next $np$ cycles, where $np$ the number of slices used, the $np$ partial outputs can be reduced into the final output activation. The user can chose any number of slices up to 16, so that \TRT can be fully utilized even with fully-connected layers of just 256 outputs. %
This cascade mode can be useful in other Deep Learning networks such as in NeuralTalk \cite{DBLP:journals/corr/KarpathyF14} where the smallest \FCLs can have 600 outputs or fewer. %

\noindent\textbf{Other Layers:} \TRT like \DaDN can process the additional layers needed by the studied networks. For this purpose the \tile includes additional hardware support for max pooling similar to \DaDN. An activation function unit is present at the output of NBout in order to apply nonlinear activations before the output neurons are written back to NM.

\subsection{SIP and Other Components}
\label{sec:computeunit}
\begin{figure}
\centering
\includegraphics[scale=0.50]{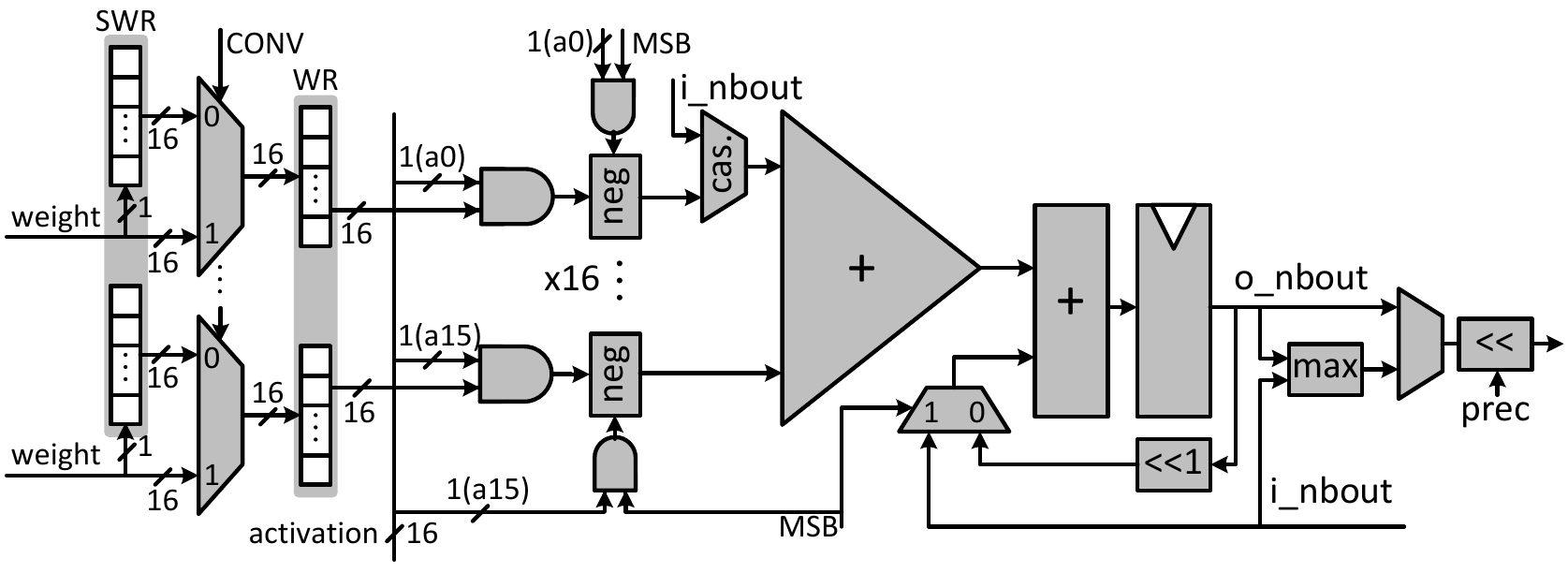}
\caption{\TRT's SIP.}
\label{fig:trt-sip}
\end{figure}

\noindent\textbf{SIP: Bit-Serial Inner-Product Units:} 
Figure \ref{fig:trt-sip} shows \TRT's Bit-Serial Inner-Product Unit (SIP). Each SIP multiplies 16 activation bits, one bit per activation, by 16 weights to produce an output activation. Each SIP has two registers, a Serial Weight Register (SWR) and a Weight Register (WR), each containing 16 16-bit subregisters. Each SWR subregister is a shift register with a single bit connection to one of the weight bus wires that is used to read weights bit-serially for FCLs. Each WR subregister can be parallel loaded from either the weight bus or the corresponding SWR subregister, to process CVLs or FCLs respectively. Each SIP includes 256 2-input AND gates that multiply the weights in the WR with the incoming activation bits, and a $16\times 16b$ adder tree that sums the partial products. A final adder plus a shifter accumulate the adder tree results into the output register OR.
In each SIP, a multiplexer at the first input of the adder tree implements the cascade mode supporting slicing the output activation computation along the SIPs of a single row. To support signed 2's complement neurons, the SIP can subtract the weight corresponding to the most significant bit (MSB) from the partial sum when the MSB is 1. This is done with negation blocks for each weight before the adder tree. Each SIP also includes a comparator (max) to support max pooling layers. 

\noindent\textbf{Dispatcher and Reducers:} Figure~\ref{fig:systems-overview}b shows an overview of the full \TRT system. As in \DaDN there is a central NM and 16 tiles. A \textit{Dispatcher} unit is tasked with reading input activations from NM always performing eDRAM-friendly wide accesses. It transposes each activation and communicates each a bit a time over the global interconnect. For CVLs the dispatcher has to maintain a pool of multiple activation bricks, each from different window, which may require fetching multiple rows from NM. However, since a new set of windows is only needed every $P_{a}^L$ cycles, the dispatcher can keep up for the layers studied. For FCLs one activation brick is sufficient. A \textit{Reducer} per title is tasked with collecting the output activations and writing them to NM. Since output activations take multiple cycles to produce, there is sufficient bandwidth to sustain all 16 tiles.

\subsection{Processing Several Activation Bits at Once}
\label{sec:twobit}
In order to improve \TRT's area and power efficiency, the number of activation bits processed at once can be adjusted at design time. The chief advantage of these designs is that less SIPs are needed in order to achieve the same throughput -- for example, processing two activation bits at once reduces the number of SIP columns from 16 to 8 and their total number to half. Although the total number of bus wires is similar, the distance they have to cover is significantly reduced. Likewise, the total number of adders required stays similar, but they are clustered closer together. A drawback of these configurations is they forgo some of the performance potential as they force the activation precisions to be multiple of the number of bits that they process per cycle. A designer can chose the configuration that best meets their area, energy efficiency and performance target.

In these configurations the weights are multiplied with several activation bits at once, and the multiplication results are partially shifted before they are inserted into their corresponding adder tree.
In order to load the weights on time, the SWR subregister has to be modified so it can load several bits in parallel, and shift that number of positions every cycle. The negation block (for 2's complement support) will operate only over the most significant product result.

\section{Evaluation}
\label{sec:evaluation}

This section evaluates \TRT's performance, energy and area compared to \DaDN . It also explores the trade-off between accuracy and performance for \TRT.
Section~\ref{sec:methodology} described the experimental methodology. Section~\ref{sec:eval:perf} reports the performance improvements with \TRT. Section~\ref{sec:eval:ee} reports energy efficiency and Section~\ref{sec:eval:area} reports \TRT's area overhead. Finally, Section~\ref{sec:eval:2b} studies a \TRT configuration that processes two activation bits per cycle.
\subsection{Methodology}
\label{sec:methodology}

\noindent\textbf{}
\DaDN, \STR and \TRT were modeled using the same methodology for consistency. A custom cycle-accurate simulator models execution time. Computation was scheduled as described by~\cite{Stripes-MICRO} to maximize energy efficiency for \DaDN. 

The logic components of the both systems were synthesized with the Synopsys Design Compiler~\cite{synopsysDC} for a TSMC 65nm library to report power and area. The circuit is clocked at 980 MHz. The NBin and NBout SRAM buffers were modelled using CACTI \cite{Muralimanoharcacti6.0:}. The eDRAM area and energy were modelled with \textit{Destiny}~\cite{destiny}. Three design corners were considered as shown in Table~\ref{tab:corners}, and the typical case was chosen for layout.

\begin{table*}[!t]
\centering
\begin{tabular}{|c|c|c|}
\hline
     & \textbf{Area overhead} &\textbf{Mean efficiency}   \\ \hline
    
Best case & 39.40\% & 0.933 \\ \hline
Typical case & 40.40\% & 1.012 \\ \hline
Worst case & 45.30\% & 1.047 \\ \hline

\end{tabular}
\caption{
Pre-layout results comparing \TRT to \DaDN. Efficiency values for FC layers. 
}
\label{tab:corners}
\end{table*}

\begin{table*}[!t]
\centering
\begin{tabular}{|c|c|c|c|c|c|c|c|c|}
\hline

 \   & \multicolumn{4}{|c|}{\textbf{Fully Connected Layers}} & \multicolumn{4}{|c|}{\textbf{Convolutional Layers}}   \\
          \cline{1-9}
\textbf{Accuracy} & \multicolumn{2}{c|}{\textbf{100\%}}&\multicolumn{2}{c|}{\textbf{99\%}} & \multicolumn{2}{c|}{\textbf{100\%}}&\multicolumn{2}{c|}{\textbf{99\%}}  \\ 
\cline{1-9}
   &  \textbf{Perf} &\textbf{Eff} & \textbf{Perf}  &\textbf{Eff} &  \textbf{Perf} &\textbf{Eff} & \textbf{Perf}  &\textbf{Eff}\\ \hline

        AlexNet & 1.61 & 1.06 & 1.80 & 1.19 & 2.32 & 1.43 & 2.52 & 1.55 \\ \hline
        VGG\_S & 1.61 & 1.05 & 1.76 & 1.16 & 1.97 & 1.21 & 1.97 & 1.21 \\ \hline
        VGG\_M & 1.61 & 1.06 & 1.77 & 1.17 & 2.18 & 1.34 & 2.29 & 1.40 \\ \hline
        VGG\_19 & 1.60 & 1.05 & 1.61 & 1.06 & 1.35 & 0.83 & 1.56 & 0.96 \\ \hline
        \textbf{geomean} & 1.61 & 1.06 & 1.73 & 1.14 & 1.91 & 1.18 & 2.05 & 1.26 \\ \hline

\end{tabular}
\caption{
Execution time and energy efficiency improvement with \TRT compared to \DaDN. 
}
\label{tab:TTT}
\end{table*}

\subsection{Execution Time} 
\label{sec:eval:perf}
Table~\ref{tab:TTT} reports \TRT's performance and energy efficiency relative to \DaDN for the precision profiles in Table~\ref{tab:TRN_precisions} separately for \FCLs, \CVLs, and the whole network. 
For the 100\% profile, where no accuracy is lost, \TRT yields, on average, a speedup of {\meanspeedupFC} over \DaDN on FCLs. With the 99\% profile, it improves to {\meanspeedupFCloss}.

There are two main reasons the ideal speedup can't be reached in practice: dispatch overhead and under-utilization. Dispatch overhead occurs on the initial $P^L_w$ cycles of execution, where the serial weight loading process prevents any useful products to be performed. In practice, this overhead is less than 2\% for any given network, although it can be as high as 6\% for the smallest layers. Underutilization can happen when the number of output neurons is not a power of two, or lower than 256. The last classifier layers of networks designed to perform recognition of ImageNet  categories \cite{imagenet} all provide 1000 output neurons, which leads to 2.3\% of the SIPs being idle.

Compared to \STR, \TRT matches its performance improvements on \CVLs while offering  performance improvements on \FCLs. We do not report the detailed results for \STR since they would have been identical to \TRT for \CVLs and within 1\% of \BASE for \FCLs. 

We have also evaluated \TRT  on\ NeuralTalk LSTM~\cite{DBLP:journals/corr/KarpathyF14} which uses long short-term memory to automatically generate image captions. Precision can be reduced down to 11 bits without affecting the accuracy of the predictions (measured as the BLEU score when compared to the ground truth) resulting in a ideal performance improvement of $1.45\times$\xspace translating into a $1.38\times$ speedup with \TRT.
 We do not include these results in Table~\ref{tab:TTT} since we did not study the \CVLs nor did we explore reducing precision further to obtain a 99\% accuracy profile.
\subsection{Energy Efficiency}
\label{sec:eval:ee}
 This section compares the \textit{Energy Efficiency} or simply efficiency of \TRT and \DaDN. Energy Efficiency is the inverse of the relative energy consumption of the two designs. As Table~\ref{tab:TTT} reports, the average efficiency improvement with \TRT across all networks and layers for the 100\% profile is {\meanefficiencyALL}. In FCLs, \TRT is more efficient than \BASE.
Overall, efficiency primarily comes from the reduction in effective computation following the use of reduced precision arithmetic for the inner product operations. Furthermore, the amount of data that has to be transmitted from the SB and the traffic between the central eDRAM and the SIPs is decreased proportionally with the chosen precision.

\begin{table*}[!t]
\centering
\begin{tabular}{|c|c|c|c|}
\hline
 & \textbf{\TRT area $(mm^2)$} &  \textbf{\TRT 2-bit area $(mm^2)$} & \textbf{\DaDN area $(mm^2)$} \\
\hline
\hline 
\textbf{Inner-Product Units}  & 57.27   (47.71\%)       & 37.66 (37.50\%)       & 17.85   (22.20\%) \\
\textbf{Synapse Buffer}       & 48.11   (40.08\%)       & 48.11 (47.90\%)       & 48.11   (59.83\%) \\
\textbf{Input Neuron Buffer}  & 3.66    (3.05\%)        & 3.66  (3.64\%)        & 3.66    (4.55\%) \\
\textbf{Output Neuron Buffer} & 3.66    (3.05\%)        & 3.66  (3.64\%)        & 3.66    (4.55\%) \\
\textbf{Neuron Memory}        & 7.13    (5.94\%)        & 7.13  (7.10\%)        & 7.13    (8.87\%) \\
\textbf{Dispatcher}           & 0.21    (0.17\%)        & 0.21  (0.21\%)        & -                \\\hline
\textbf{Total}                & 120.04 (100\%)    & 100.43 (100\%)    & 80.41 (100\%)    \\ \hline
\textbf{Normalized Total}     & $1.49\times$      & $1.25\times$     & $1.00\times$     \\ 
\hline
\end{tabular}
\caption{Area Breakdown for \TRT and \DaDN}
\label{table:TRN_area}
\end{table*}

\subsection{Area} 
\label{sec:eval:area}
Table~\ref{table:TRN_area} reports the area breakdown of \TRT and \DaDN. Over the full chip, \TRT needs $1.49\times$ the area compared to \BASE while delivering on average a \meanspeedupALL improvement in speed. Generally, performance would scale sublinearly with area for \DaDN due to underutilization. The 2-bit variant, which has a lower area overhead, is described in detail in the next section.\ %

\subsection{\texorpdfstring{$\TRT_{2b}$}{TRT2b}}
\label{sec:eval:2b}
This section evaluates the performance, energy efficiency and area for a multi-bit design as described in Section \ref{sec:twobit}, where 2 bits are processed every cycle in as half as many total SIPs. The precisions used are the same as indicated in Table~\ref{tab:TRN_precisions} for the 100\% accuracy profile rounded up to the next multiple of two. Table~\ref{tab:TBT} reports the resulting performance. The 2-bit \TRT always improves performance compared to \DaDN as the ``vs. \DaDN'' columns show. Compared to the 1-bit \TRT performance is slightly lower however given that the area of the 2-bit \TRT is much lower, this can be a good trade-off. Overall, there are two forces at work that shape performance relative to the 1-bit \TRT. There is performance potential lost due to rounding all precisions to an even number, and there is performance benefit by requiring less parallelism. The time needed to serially load the first bundle of weights is also reduced. In VGG\_19 the performance benefit due to the lower parallelism requirement outweighs the performance loss due to precision rounding. In all other cases, the reverse is true.

\begin{table*}[!t]
\centering
\begin{tabular}{|c|c|c|c|c|}
\hline

 \   & \multicolumn{2}{|c|}{\textbf{Fully Connected Layers}} & \multicolumn{2}{|c|}{\textbf{Convolutional Layers}}   \\
          \cline{1-5}
   &  \textbf{vs. DaDN} &\textbf{vs. 1b TRT}  &  \textbf{vs. DaDN} &\textbf{vs. 1b TRT} \\ %
          \cline{1-5}

        AlexNet      & +58\% & -2.06\% & +208\% &  -11.71\% \\ \hline
        VGG\_S       & +59\% & -1.25\% & +76\% &  -12.09\% \\ \hline
        VGG\_M       & +63\% &  +1.12\% & +91\% &  -13.78\% \\ \hline
        VGG\_19      & +59\% & -0.97\% & +29\% &  -4.11\% \\ \hline
\textbf{geomean} & +60\% & -0.78\% & +73\% &  -10.36\% \\ \hline

\end{tabular}
\caption{
Relative performance of 2-bit \TRT variation compared to \DaDN and 1-bit \TRT 
}
\label{tab:TBT}
\end{table*}

A hardware synthesis and layout of both \DaDN and \TRT's 2-bit variant using TSMC's 65nm typical case libraries shows that the total area overhead can be as low as 24.9\% (Table~\ref{table:TRN_area}), with an improved energy efficiency in fully connected layers of $1.24\times$ on average (Table~\ref{tab:TTT}).

\section{Related Work and Limitations}
\label{sec:related}

The recent success of Deep Learning has led to several proposals for hardware acceleration of DNNs. This section reviews some of these recent efforts. However, specialized hardware designs for neural networks is a field with a relatively long history. Relevant to \TRT, bit-serial processing hardware for neural networks has been proposed several decades ago, e.g.,~\cite{svensson1990execution,murray1988bit}. While the performance of these designs scales with precision it would be lower than that of an equivalently configured bit-parallel engine. For example, Svensson et al., uses an interesting bit-serial multiplier which requires $O(4\times p)$ cycles, where $p$ the precision in bits~\cite{svensson1990execution}. Furthermore, as semiconductor technology has progressed the number of resources that can be put on chip and the trade offs (e.g., relative speed of memory vs. transistors vs. wires) are today vastly different facilitating different designs. However, truly bit-serial processing such as that used in the aforementioned proposals needs to be revisited with today's technology constraints due to its potentially high compute density (compute bandwidth delivered per area).

In general, hardware acceleration for DNNs has recently progressed in two directions: 1)~considering more general purpose accelerators that can support additional machine learning algorithms, and 2)~considering further improvements primarily for convolutional neural networks and the two most dominant in terms of execution time layer types: convolutional and fully-connected. In the first category there are accelerators such as Cambricon~\cite{cambricon:2016} and Cambricon-X~\cite{CambriconXMICRO16}. While targeting support for more machine learning algorithms is desirable, work on further optimizing performance for specific algorithms such as \TRT is valuable and needs to be pursued as it will affect future iterations of such general purpose accelerators.

\TRT is closely related to \STRL~\cite{Stripes-CAL,Stripes-MICRO} whose execution time scales with precision but only for CVLs. \STR does not improve performance for FCLs. \TRT improves upon \STR by enabling: 1)~ performance improvements for FCLs, and 2)~slicing the activation computation across multiple SIPs thus preventing under-utilization for layers with fewer than 4K outputs. \textit{Pragmatic} uses a similar in spirit organization to \STR but its performance on CVLs depends only on the number of activation bits that are 1~\cite{Pragmatic}. It should be possible to apply the \TRT extensions to Pragmatic, however, performance in FCLs will still be dictated by weight precision. The area and energy overheads would need to be amortized by a commensurate performance improvement necessitating a dedicated evaluation study.

The \textit{Efficient Inference Engine} (EIE) uses synapse pruning, weight compression, zero activation elimination, and network retraining to drastically reduce the amount of computation and data communication when processing fully-connected layers~\cite{EIEISCA16}. An appropriately configured EIE will outperform \TRT for FCLs, provided that the network is pruned and retrained. However, the two approaches attack a different component of FCL processing and there should be synergy between them. Specifically, EIE currently does not exploit the per layer precision variability of DNNs and relies on retraining the network. It would be interesting to study how EIE would benefit from a \TRT-like compute engine where EIE's data compression and pruning is used to create vectors of weights and activations to be processed in parallel. EIE uses single-lane units whereas \TRT uses a coarser-grain lane arrangement and thus would be prone to more imbalance. A middle ground may be able to offer some performance improvement while compensating for cross-lane imbalance.

Eyeriss uses a systolic array like organization and gates off computations for zero activations~\cite{isscc_2016_chen_eyeriss} and targets primarily high-energy efficiency. An actual prototype has been built and is in full operation. Cnvlutin is a SIMD accelerator that skips on-the-fly ineffectual activations such as those that are zero or close to zero~\cite{albericio:cnvlutin}. Minerva is a DNN hardware generator which also takes advantage of zero activations and that targets high-energy efficiency~\cite{Reagen2016}. Layer fusion can further reduce off-chip communication and create additional parallelism~\cite{x-alwani-micro-fused-layer-cnn-accelerators}. As multiple layers are processed concurrently, a straightforward combination with \TRT would use the maximum of the precisions when layers are fused.

Google's Tensor Processing Unit uses quantization to represent values using 8 bits~\cite{TPUGoogle} to support TensorFlow~\cite{tensorflow2015-whitepaper}. As Table~\ref{tab:TRN_precisions} shows, some layers can use  lower than 8 bits of precision which suggests that even with quantization it may be possible to use fewer levels and to potentially benefit from an engine such as \TRT.

\subsection{Limitations} As in \DaDN this work assumed that each layer  fits on-chip. However, as networks evolve it is likely that they will increase in size thus requiring multiple \TRT nodes as was suggested in \DaDN. However, some newer networks tend to use more but smaller layers. Regardless, it would be desirable to reduce the area cost of \TRT most of which is due to the eDRAM buffers. We have not explored this possibility in this work. Proteus~\cite{ProteusICS16} is directly compatible with \TRT and can reduce memory footprint by about 60\% for both convolutional and fully-connected layers. Ideally, compression, quantization and pruning similar in spirit to EIE~\cite{EIEISCA16} would be used to reduce computation, communication and footprint. General memory compression~\cite{Mittal2016} techniques offer additional opportunities for reducing footprint and communication.

We evaluated \TRT only on CNNs for image classification. Other network architectures are important and the layer configurations and their relative importance varies. \TRT enables performance improvements for two of the most dominant layer types. We have also provided some preliminary evidence that \TRT works well for NeuralTalk LSTM~\cite{DBLP:journals/corr/KarpathyF14}. Moreover, by enabling output activation computation slicing it can accommodate relatively small layers as well. 

Applying some of the concepts that underlie the \TRT design to other more general purpose accelerators such as Cambricon~\cite{cambricon:2016} or graphics processors would certainly be more preferable than a dedicated accelerator in most application scenarios. However, these techniques are best first investigated into specific designs and then can be generalized appropriately.

We have evaluated \TRT only for inference only. Using an engine whose performance scales with precision would provide another degree of freedom for network training as well. However, \TRT needs to be modified accordingly to support all the operations necessary during training and the training algorithms need to be modified to take advantage of precision adjustments.

This section commented only on related work on digital hardware accelerators for DNNs. Advances at the algorithmic level would impact \TRT as well or may even render it obsolete. For example, work on using binary weights~\cite{courbariaux2015binaryconnect} would obviate the need for an accelerator whose performance scales with weight precision. Investigating \TRT's interaction with other network types and architectures and other machine learning algorithms is left for future work. 

\section{Conclusion}
\label{sec:theend}
This work presented \TRTL, an accelerator for inference with Convolutional Neural Networks whose performance scales inversely linearly with the number of bits used to represent values in fully-connected and convolutional layers. \TRT also enables on-the-fly accuracy vs. performance and energy efficiency trade offs and its benefits were demonstrated over a set of popular image classification networks. The new key ideas in \TRT are: 1) Supporting both the bit-parallel and the bit-serial loading of weights into processing units to facilitate the processing of either convolutional or fully-connected layers, and 2) cascading the adder trees of various subunits (SIPs) to enable slicing the output computation thus reducing or eliminating cross-lane imbalance for relatively small layers. 

\TRT opens up a new direction for research in inference and training by enabling precision adjustments to translate into performance and energy savings. These precisions adjustments can be done statically prior to execution or dynamically during execution. While we demonstrated \TRT for inference only, we believe that \TRT, especially if combined with Pragmatic, opens up a new direction for research in training as well. For systems level research and development, \TRT with its ability to trade off accuracy for performance and energy efficiency enables a new degree of adaptivity for operating systems and applications.

\bibliographystyle{ieeetr}
\bibliography{ref}

\begin{thebibliography}{10}

\bibitem{Stripes-MICRO}
P.~Judd, J.~Albericio, T.~Hetherington, T.~Aamodt, and A.~Moshovos, ``{Stripes:
  Bit-serial Deep Neural Network Computing },'' in {\em Proceedings of the 49th
  Annual IEEE/ACM International Symposium on Microarchitecture}, MICRO-49,
  2016.

\bibitem{Stripes-CAL}
P.~Judd, J.~Albericio, and A.~Moshovos, ``{Stripes: Bit-serial Deep Neural
  Network Computing },'' {\em Computer Architecture Letters}, 2016.

\bibitem{DaDiannao}
Y.~Chen, T.~Luo, S.~Liu, S.~Zhang, L.~He, J.~Wang, L.~Li, T.~Chen, Z.~Xu,
  N.~Sun, and O.~Temam, ``Dadiannao: A machine-learning supercomputer,'' in
  {\em Microarchitecture (MICRO), 2014 47th Annual IEEE/ACM International
  Symposium on}, pp.~609--622, Dec 2014.

\bibitem{Alexnet}
A.~Krizhevsky, I.~Sutskever, and G.~E. Hinton, ``Imagenet classification with
  deep convolutional neural networks,'' in {\em Advances in Neural Information
  Processing Systems 25: 26th Annual Conference on Neural Information
  Processing Systems 2012. Proceedings of a meeting held December 3-6, 2012,
  Lake Tahoe, Nevada, United States.}, pp.~1106--1114, 2012.

\bibitem{darkSilicon}
H.~Esmaeilzadeh, E.~Blem, R.~St.~Amant, K.~Sankaralingam, and D.~Burger, ``Dark
  silicon and the end of multicore scaling,'' in {\em Proceedings of the 38th
  Annual International Symposium on Computer Architecture}, ISCA '11, (New
  York, NY, USA), pp.~365--376, ACM, 2011.

\bibitem{diannao}
T.~Chen, Z.~Du, N.~Sun, J.~Wang, C.~Wu, Y.~Chen, and O.~Temam, ``{Diannao}: A
  small-footprint high-throughput accelerator for ubiquitous
  machine-learning,'' in {\em Proceedings of the 19th international conference
  on Architectural support for programming languages and operating systems},
  2014.

\bibitem{EIEISCA16}
S.~Han, X.~Liu, H.~Mao, J.~Pu, A.~Pedram, M.~A. Horowitz, and W.~J. Dally,
  ``{EIE:} efficient inference engine on compressed deep neural network,'' in
  {\em 43rd {ACM/IEEE} Annual International Symposium on Computer Architecture,
  {ISCA} 2016, Seoul, South Korea, June 18-22, 2016}, pp.~243--254, 2016.

\bibitem{albericio:cnvlutin}
J.~Albericio, P.~Judd, T.~Hetherington, T.~Aamodt, N.~E. Jerger, and
  A.~Moshovos, ``Cnvlutin: Ineffectual-neuron-free deep neural network
  computing,'' in {\em 2016 {IEEE/ACM} {International} {Conference} on
  {Computer} {Architecture} ({ISCA})}, 2016.

\bibitem{isscc_2016_chen_eyeriss}
{Chen, Yu-Hsin and Krishna, Tushar and Emer, Joel and Sze, Vivienne},
  ``{Eyeriss: An Energy-Efficient Reconfigurable Accelerator for Deep
  Convolutional Neural Networks},'' in {\em {IEEE International Solid-State
  Circuits Conference, {ISSCC} 2016, Digest of Technical Papers}},
  pp.~{262--263}, {2016}.

\bibitem{Reagen2016}
B.~Reagen, P.~Whatmough, R.~Adolf, S.~Rama, H.~Lee, S.~K. Lee, J.~M.
  Hern\'{a}ndez-Lobato, G.-Y. Wei, and D.~Brooks, ``Minerva: Enabling
  low-power, highly-accurate deep neural network accelerators,'' in {\em
  Proceedings of the 43rd International Symposium on Computer Architecture},
  ISCA '16, (Piscataway, NJ, USA), pp.~267--278, IEEE Press, 2016.

\bibitem{judd:reduced}
P.~Judd, J.~Albericio, T.~Hetherington, T.~Aamodt, N.~E. Jerger, R.~Urtasun,
  and A.~Moshovos, ``{Reduced-Precision Strategies for Bounded Memory in Deep
  Neural Nets, arXiv:1511.05236v4 [cs.LG] },'' {\em arXiv.org}, 2015.

\bibitem{kim_x1000_2014}
J.~Kim, K.~Hwang, and W.~Sung, ``X1000 real-time phoneme recognition {VLSI}
  using feed-forward deep neural networks,'' in {\em 2014 {IEEE}
  {International} {Conference} on {Acoustics}, {Speech} and {Signal}
  {Processing} ({ICASSP})}, pp.~7510--7514, May 2014.

\bibitem{caffe}
Y.~Jia, E.~Shelhamer, J.~Donahue, S.~Karayev, J.~Long, R.~Girshick,
  S.~Guadarrama, and T.~Darrell, ``Caffe: Convolutional architecture for fast
  feature embedding,'' {\em arXiv preprint arXiv:1408.5093}, 2014.

\bibitem{model-zoo}
Y.~Jia, ``Caffe model zoo,'' {\em
  https://github.com/BVLC/caffe/wiki/Model-Zoo}, 2015.

\bibitem{ProteusICS16}
P.~Judd, J.~Albericio, T.~Hetherington, T.~M. Aamodt, N.~E. Jerger, and
  A.~Moshovos, ``Proteus: Exploiting numerical precision variability in deep
  neural networks,'' in {\em Proceedings of the 2016 International Conference
  on Supercomputing}, ICS '16, (New York, NY, USA), pp.~23:1--23:12, ACM, 2016.

\bibitem{DBLP:journals/corr/KarpathyF14}
A.~Karpathy and F.~Li, ``Deep visual-semantic alignments for generating image
  descriptions,'' {\em CoRR}, vol.~abs/1412.2306, 2014.

\bibitem{synopsysDC}
{S}ynopsys, ``{D}esign {C}ompiler.''
  \url{http://www.synopsys.com/Tools/Implementation/RTLSynthesis/DesignCompiler/Pages}.

\bibitem{Muralimanoharcacti6.0:}
N.~Muralimanohar and R.~Balasubramonian, ``Cacti 6.0: A tool to understand
  large caches.''

\bibitem{destiny}
M.~Poremba, S.~Mittal, D.~Li, J.~Vetter, and Y.~Xie, ``Destiny: A tool for
  modeling emerging 3d nvm and edram caches,'' in {\em Design, Automation Test
  in Europe Conference Exhibition (DATE), 2015}, pp.~1543--1546, March 2015.

\bibitem{imagenet}
O.~Russakovsky, J.~Deng, H.~Su, J.~Krause, S.~Satheesh, S.~Ma, Z.~Huang,
  A.~Karpathy, A.~Khosla, M.~Bernstein, A.~C. Berg, and L.~Fei-Fei,
  ``{ImageNet} {Large} {Scale} {Visual} {Recognition} {Challenge},'' {\em
  arXiv:1409.0575 [cs]}, Sept. 2014.
\newblock arXiv: 1409.0575.

\bibitem{svensson1990execution}
B.~Svensson and T.~Nordstrom, ``Execution of neural network algorithms on an
  array of bit-serial processors,'' in {\em Pattern Recognition, 1990.
  Proceedings., 10th International Conference on}, vol.~2, pp.~501--505, IEEE,
  1990.

\bibitem{murray1988bit}
A.~F. Murray, A.~V. Smith, and Z.~F. Butler, ``Bit-serial neural networks,'' in
  {\em Neural Information Processing Systems}, pp.~573--583, 1988.

\bibitem{cambricon:2016}
S.~Liu, Z.~Du, J.~Tao, D.~Han, T.~Luo, Y.~Xie, Y.~Chen, and T.~Chen,
  ``Cambricon: An instruction set architecture for neural networks,'' in {\em
  2016 {IEEE/ACM} {International} {Conference} on {Computer} {Architecture}
  ({ISCA})}, 2016.

\bibitem{CambriconXMICRO16}
S.~Zhang, Z.~Du, L.~Zhang, H.~Lan, S.~Liu, L.~Li, Q.~Guo, T.~Chen, and Y.~Chen,
  ``Cambricon-x: An accelerator for sparse neural networks,'' in {\em
  Proceedings of the 49th International Symposium on Microarchitecture}, 2016.

\bibitem{Pragmatic}
J.~Albericio, P.~Judd, A.~D. Lascorz, S.~Sharify, and A.~Moshovos,
  ``Bit-pragmatic deep neural network computing,'' {\em Arxiv},
  vol.~arXiv:1610.06920 [cs.LG], 2016.

\bibitem{x-alwani-micro-fused-layer-cnn-accelerators}
M.~Alwani, H.~Chen, M.~Ferdman, and P.~Milder, ``Fused-layer cnn
  accelerators,'' in {\em 49th Annual IEEE/ACM International Symposium on
  Microarchitecture (MICRO)}, 2016.

\bibitem{TPUGoogle}
N.~Jouppi, ``{Google supercharges machine learning tasks with TPU custom
  chip}.''
  \url{https://cloudplatform.googleblog.com/2016/05/Google-supercharges-machine-learning-tasks-with-custom-chip.html},
  2016.
\newblock [Online; accessed 3-Nov-2016].

\bibitem{tensorflow2015-whitepaper}
M.~Abadi, A.~Agarwal, P.~Barham, E.~Brevdo, Z.~Chen, C.~Citro, G.~S. Corrado,
  A.~Davis, J.~Dean, M.~Devin, S.~Ghemawat, I.~Goodfellow, A.~Harp, G.~Irving,
  M.~Isard, Y.~Jia, R.~Jozefowicz, L.~Kaiser, M.~Kudlur, J.~Levenberg,
  D.~Man\'{e}, R.~Monga, S.~Moore, D.~Murray, C.~Olah, M.~Schuster, J.~Shlens,
  B.~Steiner, I.~Sutskever, K.~Talwar, P.~Tucker, V.~Vanhoucke, V.~Vasudevan,
  F.~Vi\'{e}gas, O.~Vinyals, P.~Warden, M.~Wattenberg, M.~Wicke, Y.~Yu, and
  X.~Zheng, ``{TensorFlow}: Large-scale machine learning on heterogeneous
  systems,'' 2015.
\newblock Software available from tensorflow.org.

\bibitem{Mittal2016}
S.~Mittal and J.~S. Vetter, ``A survey of architectural approaches for data
  compression in cache and main memory systems,'' {\em IEEE Trans. Parallel
  Distrib. Syst.}, vol.~27, pp.~1524--1536, May 2016.

\bibitem{courbariaux2015binaryconnect}
M.~Courbariaux, Y.~Bengio, and J.-P. David, ``Binaryconnect: Training deep
  neural networks with binary weights during propagations,'' in {\em Advances
  in Neural Information Processing Systems}, pp.~3123--3131, 2015.

\end{thebibliography}

\end{document}